\documentclass{article}

\usepackage[preprint]{corl_2021} % Uncomment for pre-prints (e.g., arxiv); This is like ``final'', but will remove the CORL footnote.

\usepackage{xcolor}
\usepackage{graphicx}
\usepackage{amsmath}
\usepackage{booktabs}
\usepackage{caption}
\usepackage{subcaption}
\usepackage{tikz}
\usepackage{pgfplots}
\usetikzlibrary{backgrounds}
\pgfplotsset{compat=1.17}
\usepgfplotslibrary{colorbrewer}
\pgfplotsset{cycle list/Set1}
\usepackage{multirow}
\usepackage[inline]{enumitem}

\usepackage[normalem]{ulem} % for \sout

\newlist{inline-enum}{enumerate*}{1}
\setlist[inline-enum]{label=(\arabic*)}

\newcommand{\remark}[3]{{\color[hsb]{#2,1.0,0.7} [#1: #3]}}
\newcommand{\yahav}[1]{\remark{Yahav}{0.3}{#1}}
\newcommand{\jeff}[1]{\remark{Jeff}{0.6}{#1}}
\newcommand{\ken}[1]{\remark{Ken}{0.0}{#1}}

\newcommand{\algname}{Dex-NeRF}
\newcommand{\NEW}[1]{{\color{blue}#1}}
\newcommand{\OLD}[1]{\sout{#1}}

\renewcommand{\NEW}[1]{#1}
\renewcommand{\OLD}[1]{}

\title{\algname{}: Using a Neural Radiance Field \\ to Grasp Transparent Objects}

\author{
  Jeffrey Ichnowski \thanks{Equal contribution.} \\
  The AUTOLAB \\
  University of California Berkeley \\
  United States \\
  \texttt{jeffi@berkeley.edu}
   \And
   Yahav Avigal \footnotemark[1] \\
   The AUTOLAB \\
   University of California Berkeley \\
   United States \\
   \texttt{yahav\_avigal@berkeley.edu} \\
   \AND
   Justin Kerr \\
   The AUTOLAB \\
   University of California Berkeley \\
   United States \\
   \texttt{justin\_kerr@berkeley.edu}\\
   \And
   Ken Goldberg \\
   The AUTOLAB \\
   University of California Berkeley \\
   United States \\
   \texttt{goldberg@berkeley.edu}\\
}

\begin{document}
\maketitle

\begin{abstract}
    The ability to grasp and manipulate transparent objects is a major challenge for robots. Existing depth cameras have difficulty detecting, localizing, and inferring the geometry of such objects. We propose using neural radiance fields (NeRF) to detect, localize, and infer the geometry of transparent objects with sufficient accuracy to find and grasp them securely. We leverage NeRF's view-independent learned density, place lights to increase specular reflections, and perform a transparency-aware depth-rendering that we feed into the Dex-Net grasp planner. We show how additional lights create specular reflections that improve the quality of the depth map, and test a setup for a robot workcell equipped with an array of cameras to perform transparent object manipulation. We also create synthetic and real datasets of transparent objects in real-world settings, including \NEW{singulated objects,} cluttered tables, \OLD{a laboratory setting,} and the top rack of a dishwasher. In each setting we show that NeRF and Dex-Net are able to reliably compute robust grasps on transparent objects\NEW{, achieving 90\,\% and 100\,\% grasp success rates in physical experiments on an ABB YuMi, on objects where baseline methods fail}.  
%
%
% Do not include the next line in the abstract upload!
%
See \url{https://sites.google.com/view/dex-nerf} for code, video, and datasets.
\end{abstract}

\section{Introduction}
\vspace{-12pt}
Transparent objects are common in homes, restaurants, retail packaging, \OLD{bio }labs, gift shops, hospitals, and industrial warehouses.
Effectively automating robotic manipulation of transparent objects could have a broad impact, from helping in everyday tasks and performing tasks in hazardous environments. Existing depth cameras assume that surfaces of observed objects \OLD{are Lambertian , }reflect\OLD{ing} light uniformly in all directions, but this assumption does not hold for transparent objects as their appearance varies significantly under different \NEW{view directions and} illumination conditions due to reflection and refraction\OLD{ properties of transparent materials}.
In this paper, we propose and demonstrate \NEW{\emph{\algname{}},} a new method to sense the geometry of transparent objects and allow for robots to interact with them---potentially enabling automation of new tasks.

\begin{figure*}[t]
    \centering % trim = left bottom right top
    %\newlength{\myl}
    %\settodepth{\myl}{% Table w/o pipeline is is 282.285 x (95.785 + 90.785) pt
    \begin{tabular}{@{}l@{\hspace{4pt}}r@{}}
      \begin{tabular}{@{}c@{\hspace{4pt}}c@{}}
         \includegraphics[height=90pt,trim=240 0 260 0,clip]{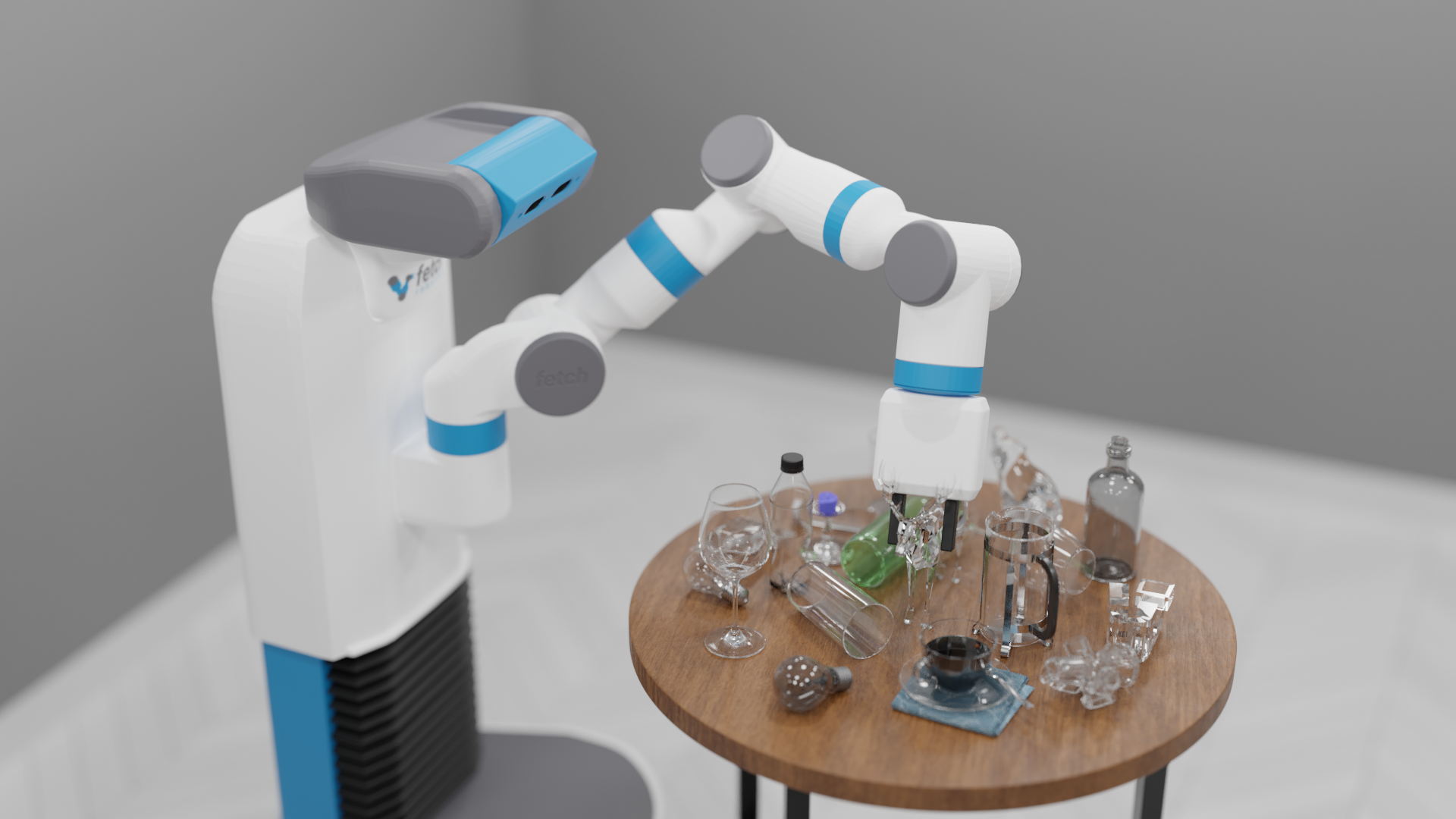} &
         \includegraphics[height=90pt]{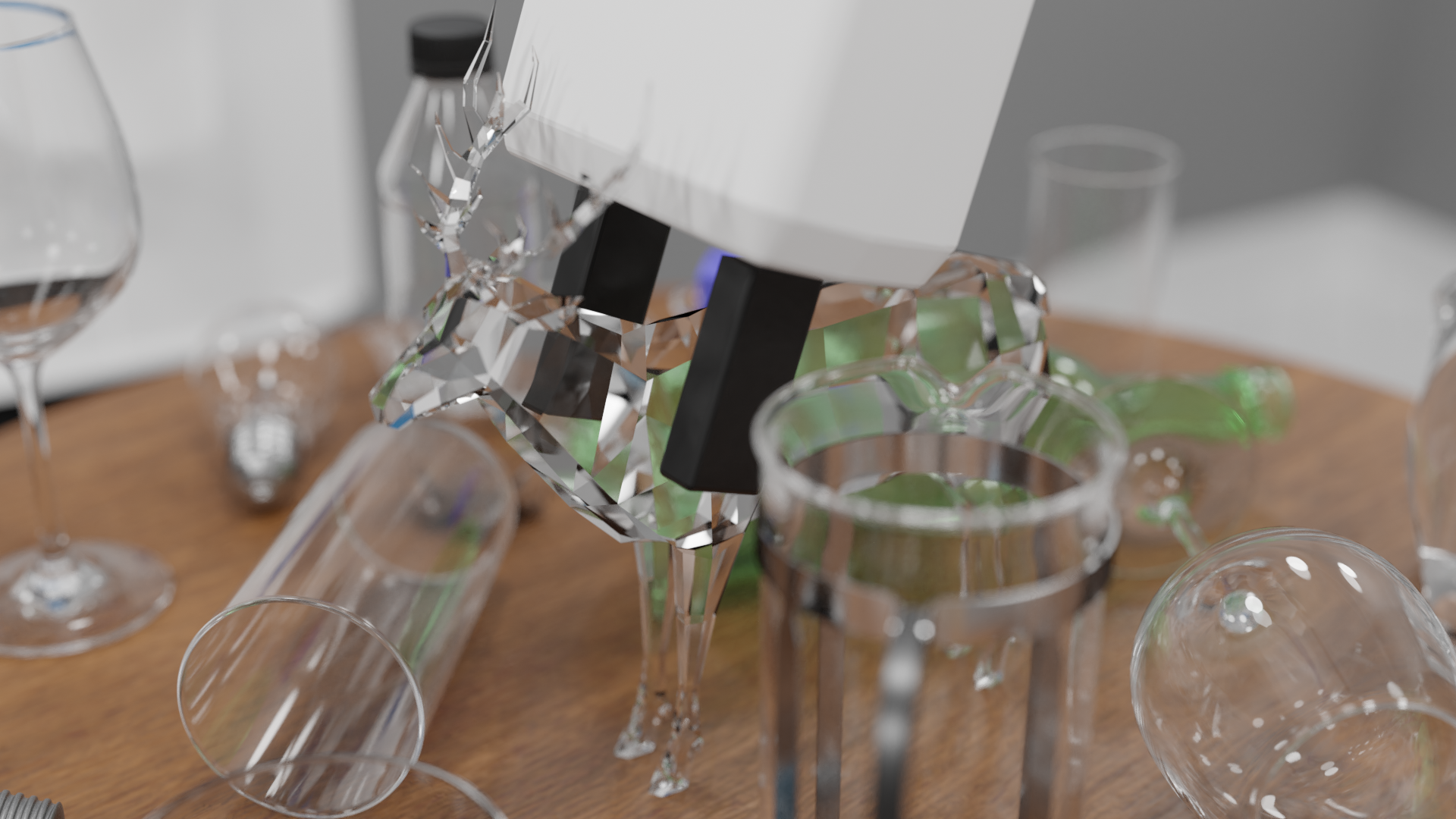} \\
         \includegraphics[height=90pt,trim=20 0 90 0,clip]{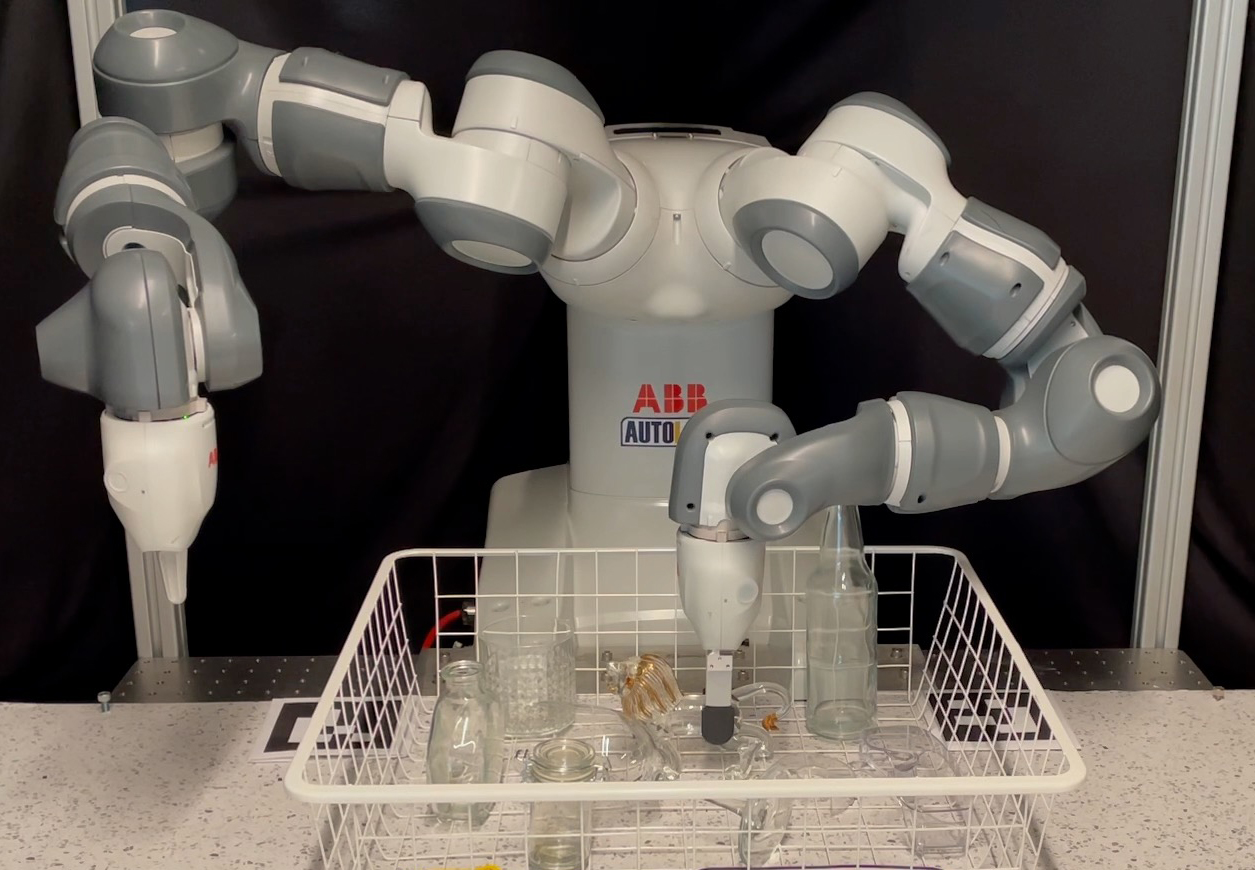} &
         \includegraphics[height=90pt]{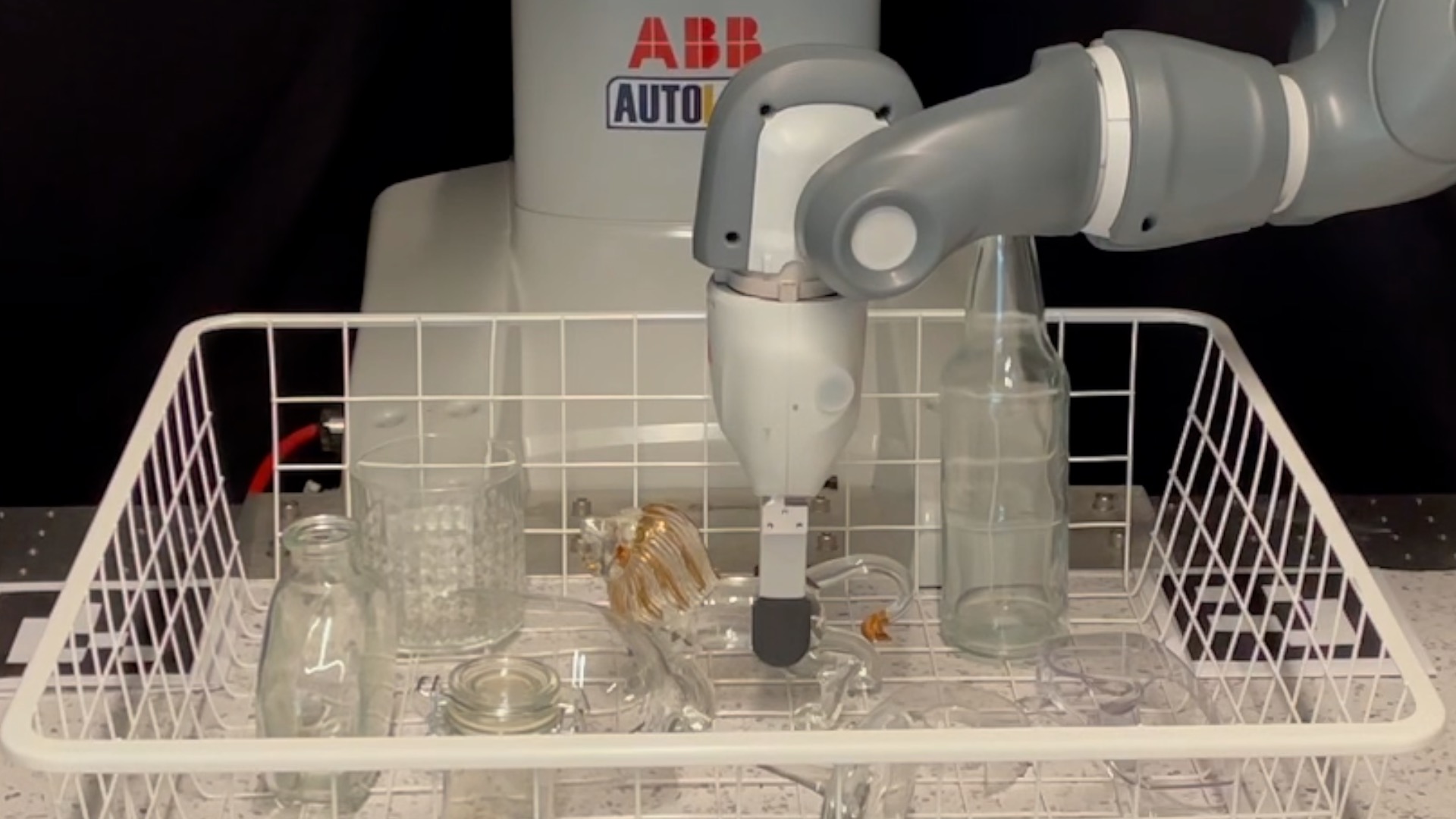}
      \end{tabular} & %} Width = \the\myl
      \raisebox{-.48\totalheight}{\includegraphics[]{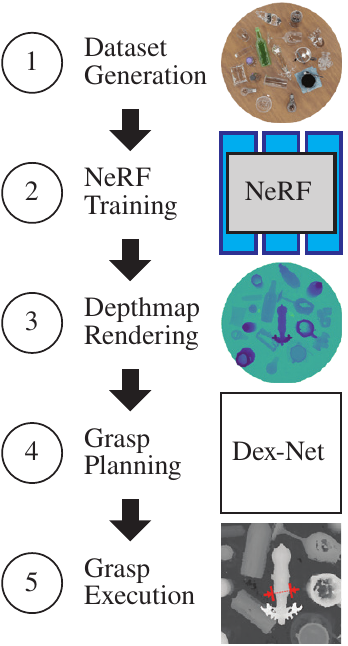}} % <-- image is 100pt wide, and does not need scaling.
    \end{tabular}
    \vspace{-6pt}
    \caption{\textbf{Using NeRF to grasp transparent objects}
    \NEW{Given a scene with transparent objects (left column), we the pipeline on the right to compute grasps (middle column).
%    Given a scene with transparent objects (top left), we (bottom row) train a NeRF model, compute a transparency-aware depth rendering, compute a grasp with Dex-Net, and then execute the grasp.  The quality of the depth map is high enough that the grasp can be executed without unwanted collisions (top right). \NEW{
The top row shows \algname{} working in a simulated scene while the bottom row shows it working in a physical scene.}}
    \label{fig:pipeline}
    \vspace{-4pt}
\end{figure*}

\NEW{%
\algname{} uses a Neural Radiance Fields (NeRF) as part of a pipeline (Fig.~\ref{fig:pipeline}, right) to compute and execute robots grasps on transparent objects.}\OLD{
The method builds on Neural Radiance Fields (NeRF).} %
While NeRF was originally proposed as an alternative for explicit volumetric representations \NEW{to}\OLD{, and shown to realistically} render novel views of complex scenes~\cite{mildenhall2020nerf}, it can also reconstruct the geometry of the scene. In particular, due to the view-dependent nature of the NeRF model, it can learn to accurately represent the geometry associated with transparency. The only input requirement to train a NeRF model is a set of images taken from a camera with known intrinsics (e.g., focal length, distortion) and extrinsics (position and orientation in the world). While the intrinsics can be determined from calibration techniques, and/or from the camera itself, determining the extrinsics is often a challenge~\cite{schoenberger2016sfm,schoenberger2016mvs}; however, robots in a fixed workcell, or using a camera attached to a manipulator arm with accurate encoders, can readily determine camera extrinsics. This makes NeRF a particularly good match for robot manipulators.

\OLD{While various grasp-planning methods use different input geometry to compute grasps, e.g., depth maps~\cite{mahler2017dex}, point clouds~\cite{xu20206dof}, and octrees~\cite{avigal2021avplug}, in this paper we propose a method to render a high-quality depth map from a NeRF model to then pass to Dex-Net~\cite{mahler2017dex} to compute a grasp.  While standard depth cameras have gaps in their depth information that needs to be processed out with hole-filling techniques, the depth map rendering from NeRF is directly usable.  We observe that other grasp-planning techniques may similarly be able to plan grasps from NeRF models as well.}

%In this paper we will show how to use it for transparent object recognition and manipulation. Stay tuned...
In experiments, we show qualitatively and quantitatively that NeRF-based grasp-planning can achieve high accuracy on NeRF models trained from photo-realistic synthetic images and from real images\NEW{, and achieve 90\,\% or better grasp success rates on real objects}.

The contributions of this paper are:
\begin{inline-enum}
    \item Integration of NeRF with robot grasp planning,
    \item A transparency-aware depth rendering method for NeRF,
    \item Experiments on synthetic and real images showing NeRF with Dex-Net generates high-quality grasps,
    \item Synthetic and real image datasets with transparent objects for training NeRF models.
\end{inline-enum}
\section{Related Work}
\vspace{-8pt}
\paragraph{Detecting Transparent Objects}
%\vspace{-4pt}
For robots to interact with transparent objects, they must first be able to detect them. The most recent approaches \OLD{to solve the problem of }detecting and recognizing transparent objects are data-driven. Lai et al.~\cite{lai2015transparent}, then Khaing et al.~\cite{khaing2018transparent}, proposed using a Convolutional Neural Network (CNN) to detect transparent objects in RGB images. Recently, Xie et al.~\cite{xie2021segmenting} developed a transformer-based pipeline~\cite{vaswani2017attention} used for transparent object segmentation. Other methods rely on deep-learning models to predict the object pose. Phillips et al.~\cite{phillips2016seeing} trained a random forest to detect the contours of transparent objects for the purpose of pose estimation and shape recovery. Xu et al.~\cite{xu20206dof} proposed a two-stage method for estimating the 6-degrees-of-freedom (DOF) pose of a transparent object with a single RGBD image by replacing the noisy depth values with estimated values and training a DenseFusion-like network structure~\cite{wang2019densefusion} to predict the object's 6-DOF pose. Sajjan et al.~\cite{sajjan2020clear} \NEW{extend this}\OLD{take it a step further} and incorporate a neural network trained for 3D pose estimation of transparent objects in a robotic picking pipeline, while Zhou et al.~\cite{zhou2019glassloc, zhou2020lit} train a grasp planner directly on raw images from a light-field camera. Zhu et al.~\cite{zhu2021rgb} used an implicit function to complete missing depth given noisy RGBD observation of transparent objects.
However, these data-driven methods rely on large annotated datasets that are hard to curate, whereas \algname{} does not require any prior \OLD{curated }dataset.

\paragraph{Neural Radiance Fields}
%\vspace{-4pt}
Recently, implicit neural representations have led to significant progress in 3D object shape representation~\cite{mescheder2019occupancy, park2019deepsdf, chen2019learning} and encoding the geometry and appearance of 3D scenes~\cite{sitzmann2019scene, mildenhall2020nerf}. Mildenhall et al.~\cite{mildenhall2020nerf} presented Neural Radiance Fields (NeRF), a neural network whose input is a 3D coordinate with an associated view direction, and output is the volume density and view-dependent emitted radiance at
that coordinate. Due to its view-dependent emitted radiance prediction, NeRF can be used to represent non-Lambertian effects such as specularities and reflections, and therefore capture the geometry of transparent objects. However, NeRF is slow to train and has low data efficiency. Yu et al.~\cite{yu2021plenoctrees} proposed \emph{Plenoctrees}, mapping coordinates to spherical harmonic coefficients, shifting the view-dependency from the input to the output. In addition, Plenoctrees pre-samples the model into a sparse octree structure, achieving a significant speedup in training over NeRF. \OLD{Jain et al.~\cite{jain2021putting} showed how adding an auxiliary semantic consistency loss to the training procedure can encourage realistic renderings at novel poses using significantly fewer views during training.} Deng et al.~\cite{jaxnerf} proposed JaxNeRF, an efficient JAX implementation of NeRF that was able to reduce the training time of a NeRF model from over a day to several hours. \NEW{Deng et al.~\cite{deng2021depth} add depth supervision to train NeRF 2 to 6$\times$ faster given fewer training views.} In this work, we propose to use NeRF to recover the \OLD{shape and }geometry of transparent objects for the purpose of robotic manipulation.

\paragraph{Robotic Grasping}
%\vspace{-4pt}
Traditional robot grasping methods analyze the object shape to identify successful grasp poses~\cite{kleeberger2020survey, bicchi2000robotic, murray2017mathematical}. Data-driven approaches learn a prior using labeled data~\cite{kappler2015leveraging, prattichizzo2008springer} or through self-supervision over many trials in a simulated or physical environment~\cite{jang2018grasp2vec, peng2021self} and generalize to grasping novel objects with unknown geometry.
Both approaches rely on RGB and depth sensors to generate a sufficiently accurate observation of the target object surface, such as depth maps~\cite{mahler2017dex, lenz2015deep, redmon2015real}, point clouds~\cite{mousavian20196, qin2020s4g, sundermeyer2021contact,xu20206dof}, octrees~\cite{avigal2021avplug}, or a truncated signed distance function (TSDF)~\cite{breyer2021volumetric, song2020grasping} from which it can compute the grasp pose. 
\NEW{While various grasp-planning methods use different input geometry to compute grasps, %e.g., depth maps~\cite{mahler2017dex}, point clouds~\cite{xu20206dof}, and octrees~\cite{avigal2021avplug}, 
in this paper we propose a method to render a high-quality depth map from a NeRF model to then pass to Dex-Net~\cite{mahler2017dex} to compute a grasp.  While standard depth cameras have gaps in their depth information that needs to be processed out with hole-filling techniques, the depth map rendering from NeRF is directly usable. It is possible that other grasp-planning techniques may be able to plan grasps from NeRF models.}

%Yahav~\cite{rivera2010extending}\yahav{what is this??????}
\section{Problem Statement}
\vspace{-4pt}
 We assume an environment that has an array of cameras at fixed known locations, or that the robot can manipulate a camera (e.g., wrist-mounted) to capture multiple images of the scene. Given the environment contains rigid transparent objects, compute a frame for a robot gripper that will result in a stable grasp of a transparent object. 
% Second, we assume the stability of grasps is independent of the objects transparency---in other words, if a grasp planner is given the geometry of the scene, the quality of the grasps it generates does not depend on the transparency of the object.  We make this second assumption, observing that transparent objects from example robot automation scenarios are made from glass or plastic and thus have rigidity consistent with the training models for grasp planners.
\section{Method}
\vspace{-4pt}
In this section, we provide a brief background on NeRF, then describe recovering geometry of transparent objects, integrating with grasp analysis, and improving performance with additional lights.

\subsection{Preliminary: Training NeRF}
\vspace{-4pt}
\OLD{The }NeRF\OLD{ model}~\cite{mildenhall2020nerf} learns a neural scene representation that maps a \OLD{single continuous }5D coordinate containing a spatial location $(x, y, z)$ and viewing direction $(\theta, \phi)$ to the volume density $\sigma$ and RGB color $\mathbf{c}$\OLD{ at that spatial location}. Training \NEW{NeRF's}\OLD{is performed using a} multilayer perceptron (MLP) \OLD{and }requires multi-view RGB images of a static scene with their corresponding camera poses and intrinsic parameters\OLD{, as well as the scene bounds}. The expected color $C(\mathbf{r})$ of the camera ray $\mathbf{r} = \mathbf{o} + t\mathbf{d}$ between near and far scene bounds $t_n$ and $t_f$ is:
\begin{equation}
    \label{gt_color}
    C(\mathbf{r}) = \int_{t_n}^{t_f}T(t)\sigma(\mathbf{r}(t))\mathbf{c}(\mathbf{r}(t),\mathbf{d})dt,
\end{equation}
where $T(t) = \exp\left(-\int_{t_n}^{t}\sigma(\mathbf{r}(s))ds \right)$ is the probability that the camera ray travels from near bound $t_n$ to point $t$ without hitting any surface.
NeRF approximates the expected color $\hat{C}(\mathbf{r})$ as:
\begin{equation}
\hat{C}(\mathbf{r}) = \sum^N_{i=1}{T_i(1-\exp(-\sigma_i \delta_i))\mathbf{c}_i},
\end{equation}
where $T_i = \exp\left(-\sum^{i-1}_{j=1}\sigma_j \delta_j\right)$ and $\delta_i = t_{i+1}-t_i$ is the distance between consecutive samples on the ray $\mathbf{r}$. During training, NeRF minimizes the error between the rendered and ground truth rays' colors using gradient descent.

%As can be observed
\OLD{%
From Eq.~\ref{gt_color}, the RGB color $\mathbf{c}(\mathbf{x},\mathbf{d})$ is view-dependent while the volume density $\sigma(\mathbf{x})$ is view-independent. This allows NeRF to represent non-Lambertian effects, such as reflections and refractions, and at the same time account for the occupancy at a spatial location. %, as described in Fig.~\ref{fig:figurine_mesh}.

We exploit this property of NeRF to estimate the position of transparent surfaces that may be invisible from some view but visible from another, allowing NeRF to detect the transparent surface via the view-independent volume density. %
}
\begin{figure}[t]
    \centering
    % trim = left bottom right top
    % \textwidth=5.5in * 72.27pt/in = 397.485 pt
    % 397.485 - 4pt*2 = 389.485 / 3 = 129.8pt
    \footnotesize
    \begin{tabular}{@{}c@{\hspace{4pt}}c@{\hspace{2pt}}c@{}}
        \includegraphics[width=129pt]{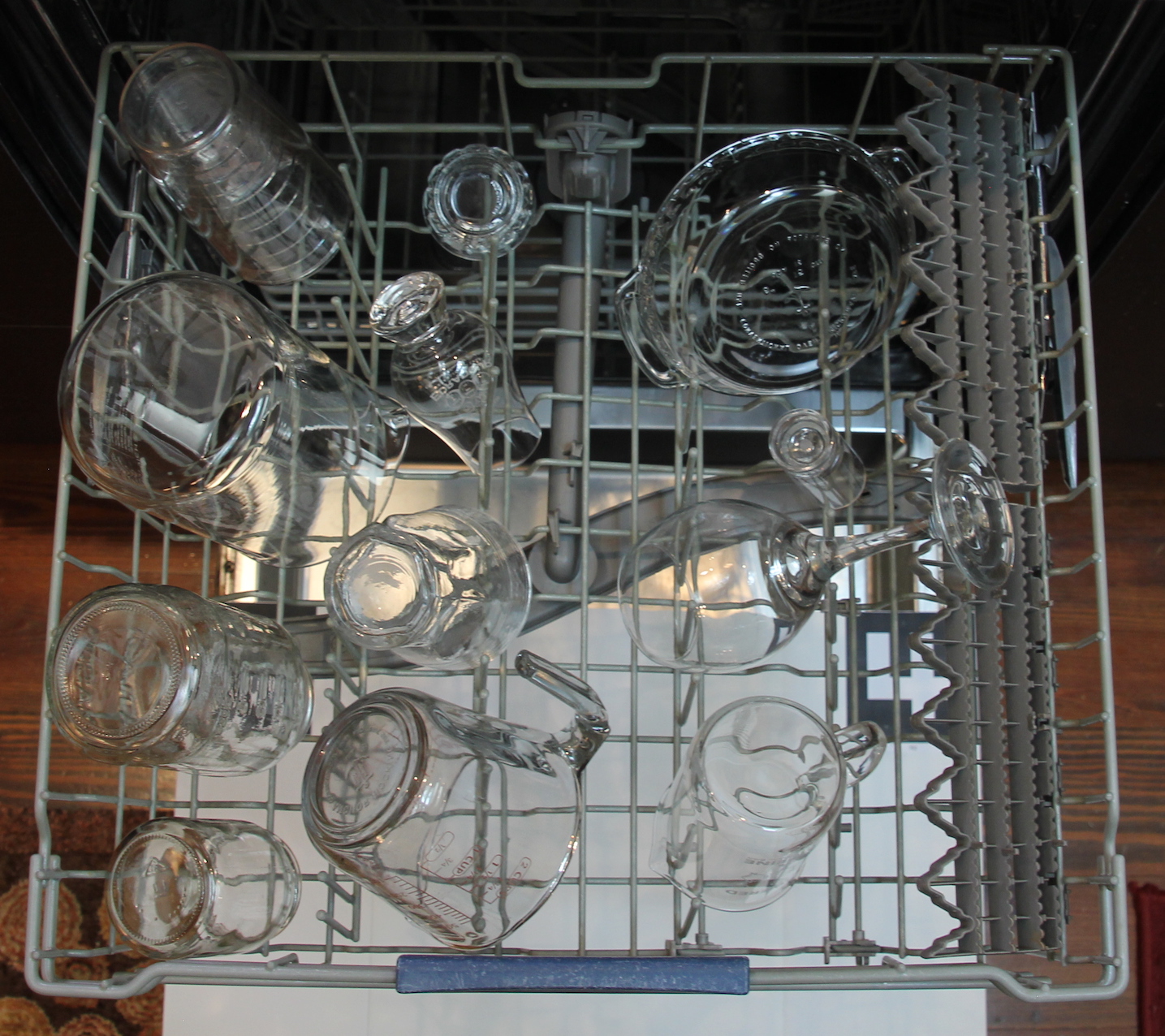} & %
        \includegraphics[width=129pt,trim=294 4 186 4,clip]{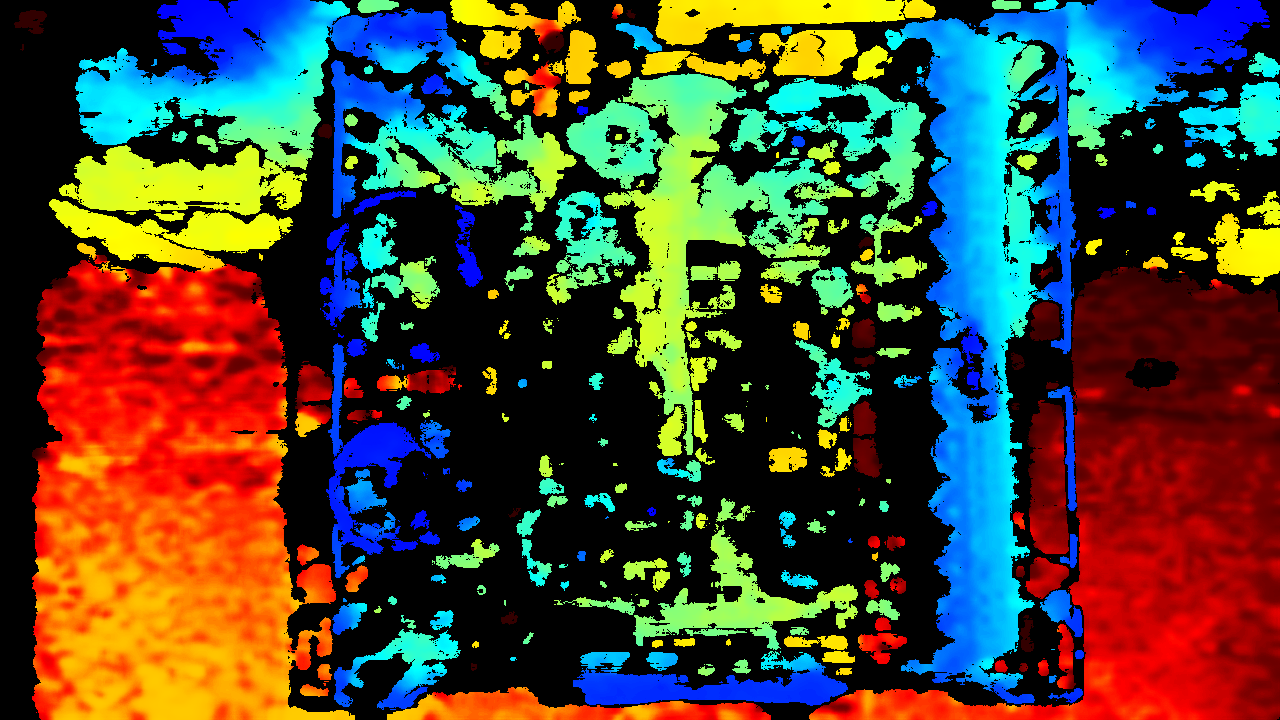} & %
        \includegraphics[width=129pt,trim=0 23 0 0,clip]{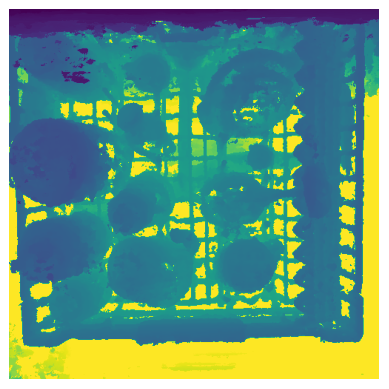} \\
        % \includegraphics[width=129pt]{images/table_scene_1_real.png} &
        % \includegraphics[width=129pt,trim=310 120 440 60,clip]{images/realsense-table_scene_1_2_Depth.png} &
        % \includegraphics[trim=70 0 70 0,clip,origin=c,angle=-90,width=129pt]{images/table_scene_1_ours.png} \\
        % \midrule
        % Real Image & RealSense Depth & Depth (Ours) \\
        % \bottomrule
        Real Image & RealSense Depth & Depth (\algname{})
    \end{tabular}\vspace{-4pt}
    \caption{\footnotesize{{\bf Comparison to RealSense Depth Camera}. We compare the results of the proposed pipeline in a real-world setting against the depth map produced by an Intel RealSense camera.  In the left image is the real-world scene, the middle shows the depth image from the RealSense, and the right shows the result of our pipeline.  The color scheme in the RealSense image is provided by the RealSense SDK, while the color scheme in the right column is from MatPlotLib.  We observe that the RealSense depth camera is unable to recover depth from a large portion of the scene, shown in black.  On the other hand, the proposed pipeline, while having a few holes, can recover depth for most of the scene.}}
    \label{fig:nerf_vs_realsense}
    \vspace{-6pt}
\end{figure}

\subsection{Recovering Geometry of Transparent Objects}
\vspace{-4pt}
\begin{figure}[t]
    \centering
    \includegraphics[width=\linewidth]{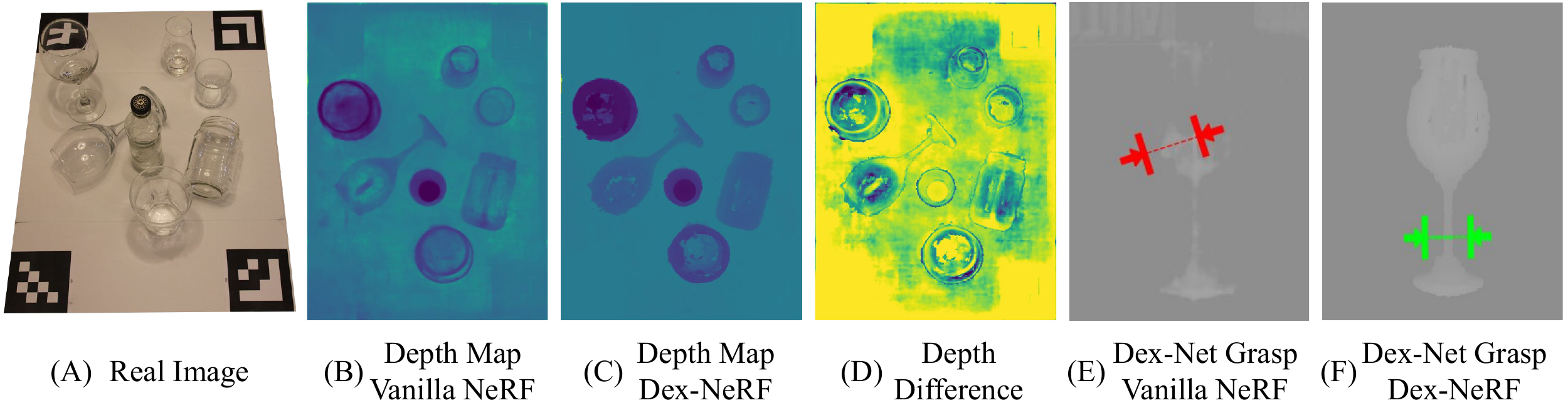} %
    \vspace{-12pt}
    \caption{ \NEW{\textbf{Using NeRF to render depth for grasping transparent objects}. \algname{} uses a transparency-aware depth rendering to render depth maps that can be used for grasp planning. In contrast, Vanilla-NeRF's depth maps are filled with holes and result in poor grasp predictions.}} %
    % \caption{\textbf{Depth sensing in the real-world.}  RGBD sensors, such as the RealSense use structured light and disparity to compute a depth image, and struggle with transparent objects.  On the other hand, NeRF is able to learn the geometry of transparent objects, and with appropriate depth rendering can produce a high-quality depth map that includes transparent object surfaces.
    % The top row shows a view from a Cannon EOS dataset, while the bottom row shows a view from a RealSense dataset (see Sec.~\ref{sec:experiments} for description).}
    \label{fig:depth_difference} %
    \vspace{-12pt}
\end{figure}
\NEW{%
We observe that NeRF does not directly support transparent object effects---it casts a single ray per source image pixel without reflection, splitting, or bouncing.
The incorporation of the viewing-direction in its regression and supervising with view-dependent emitted radiance allows recovery of non-Lambertian effects such as reflections from a specular surface. However, while RGB color $\mathbf{c}$ is view-dependent, the volume density $\sigma$ is not---meaning NeRF has to learn a non-zero $\sigma$ to represent any color at that spatial location.  Visually, the usual result is that the transparent object shows up as a “ghostly” or “blurry” version of the original object.}

When training, a NeRF model learns a density $\sigma$ of each spatial location.  This density corresponds to the transparency of the point, and serves to help learn how much a spatial location contributes to the color of a ray cast through it. \NEW{Although NeRF converts each $\sigma_i$ to an occupancy probability $\alpha_i = 1 - \exp(-\sigma_i \delta_i)$, where $\delta_i$ is the distance between integration times along the ray, thus implicitly giving $\alpha_i$ an upper bound of 1, it does not place a bound on the raw $\sigma$ value. We use the raw value} of $\sigma$ to determine if a point in space is occupied.

\OLD{In the NeRF model, $\sigma$ is view-independent, a natural choice for scenes that contain non-transparent objects, since, while color contributions are view-dependent, the occupancy of points in such a scene is not.  When training NeRF for a scene that contains transparent objects, one may consider updating $\sigma$ to be view-dependent, under the notion that this would properly capture the effect of specular reflections---which are so bright as to obscure the visibility of anything behind them.
While a view-dependent $\sigma$ may help with view-dependent rendering of scenes with transparent objects, we hypothesize that a view-dependent $\sigma$, would make recovering the geometry of transparent objects more difficult.  With a view-independent $\sigma$, we only have to query a single point in space to determine if it is non-transparent from some angle, whereas a view-dependent $\sigma$ would have to check all view direction for that same point in space, as a specular reflection that reveals the occupancy may only be visible from a few view directions.

The downside of view-independent $\sigma$ is that the RGB reconstruction of transparent objects is inaccurate, appearing smudged and ghostly.  This appears to be the result of the network learning a $\sigma$ value between transparent and opaque---transparent for views which see through the point, and opaque for views that see specular reflections at the point.}

%\jeff{Perhaps do an ablation on view-dependent $\sigma$ in supplemental materials?}
%\jeff{Discuss other sources of non-transparency, e.g. tangential views, inperfections, dirt.  Also, maybe discuss refraction?}

\subsection{Rendering Depth for Grasp Analysis}
\vspace{-4pt}
We propose using Dex-Net to compute grasps on transparent objects.  Dex-Net computes candidate grasp poses given a depth image of a scene. To generate a depth image, we consider two candidate reconstructions of depth.
First, we use the same rendering of sampled points along a camera ray that NeRF uses.  This \emph{Vanilla NeRF} reconstruction first converts $\sigma_i$ to an occupancy probability \NEW{$\alpha_i$}\OLD{$\alpha_i = 1 - \exp\left(-\sigma_i \delta_i \right)$ by multiplying the distance between adjacent samples $\delta_i = t_{i+1} - t_i$}.  It then applies the transformation $w_i = \alpha_i\prod_{j=1}^{i-1}\left(1-\alpha_j\right)$.  To render depth at pixel coordinate $[u,v]$, it computes the sum of sample distances from the camera weighted by the termination probability $D[u,v] = \sum_{i=1}^N w_i \delta_i$.
%First, the same rendering of camera rays that NeRF applies:
%the following transformations to the volume density $\sigma$:
% \begin{align}
%     \label{alpha}
%     \alpha_i &= 1 - \exp\left(-\sigma_i \delta_i \right) \\
%     \label{weights}
%     w_i &= \alpha_i\prod_{j=1}^{i-1}\left(1-\alpha_j\right) \\
%     \label{depth}
%     D[u,v] &= \sum_{i=1}^N w_i \delta_i.
% \end{align}
% This converts $\sigma_i$ into an occupancy probability $\alpha_i$ by multiplying by the distance between adjacent samples $\delta_i = t_{i+1} - t_i$ and applies the transformation in Eq.~\ref{alpha}. To render depth, it then computes the ray termination probability at each sample $w_i$ in Eq.~\ref{weights}. The depth at pixel coordinate $[u,v]$ is computed as the sum of the samples distances from the camera weighted by the termination probability, as described in Eq.~\ref{depth}. 
When applied on transparent objects, however, this results in noisy depth maps, as shown in Fig.~\ref{fig:depth_difference}.

\begin{figure}[t]
    \centering
    \begin{tabular}{c@{\hspace{5pt}}c@{\hspace{5pt}}c@{\hspace{5pt}}c@{\hspace{5pt}}c@{\hspace{5pt}}}
    {\includegraphics[width=.18\linewidth]{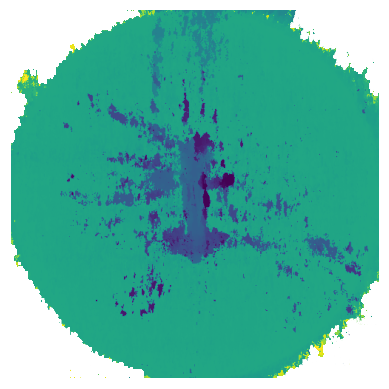}}&{\includegraphics[width=.18\linewidth]{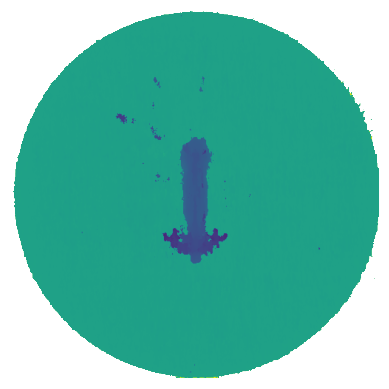}}&{\includegraphics[width=.18\linewidth]{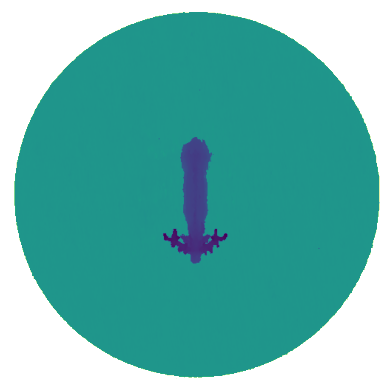}}&{\includegraphics[width=.18\linewidth]{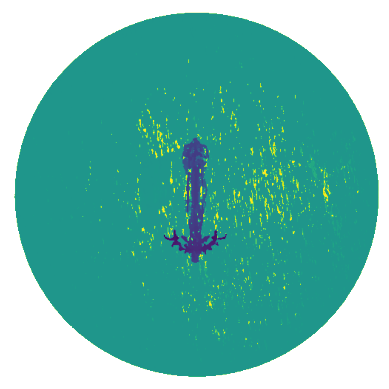}}&{\includegraphics[width=.18\linewidth]{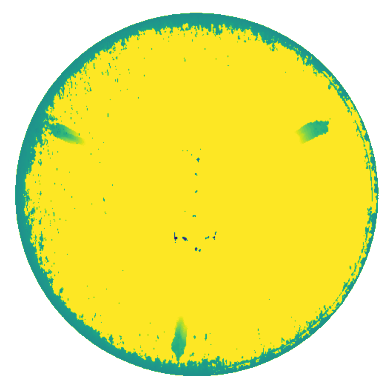}} \\
    $\sigma=1$&$\sigma=5$&$\sigma=15$&$\sigma=150$&$\sigma=500$ %\\ 
    %{\includegraphics[width=.18\linewidth]{images/cam_grid_9.png}}&{\includegraphics[width=.18\linewidth]{images/cam_grid_16.png}}&{\includegraphics[width=.18\linewidth]{images/cam_grid_25.png}}&{\includegraphics[width=.18\linewidth]{images/cam_grid_36.png}}&{\includegraphics[width=.18\linewidth]{images/cam_grid_49.png}} \\
    %$9$ Cameras&$16$ Cameras&$25$ Cameras&$36$ Cameras&$49$ Cameras
    \end{tabular}
    \caption{\footnotesize{{\bf depth rendering using NeRF with different thresholds}  Here we show the effect of the threshold value on the depth rendering on an isolated deer figurine.  Values too low result in excess noise, while values too high cause parts of the scene to disappear.}}
    \label{fig:sigma}
\end{figure}

Instead, we consider a second, transparency-aware method that searches for the first sample along the ray for which $\sigma_i > m$ where $m$ is a fixed threshold. The depth is then set to the distance of that sample $\delta_i$. We explore different values for $m$, and observe that low values result in a noisy depth map while high values create holes in the depth map. In our experiments we set $m=15$ (see Fig.~\ref{fig:sigma}). 

% \jeff{This would be a good place to explain the threshold value.}

\subsection{Improving Reconstruction with Light Placement} % {Optimizing Reflections}
\vspace{-4pt}
%Scene Lights Setup - positioning the lights and cameras to maximize the observed reflections 
For NeRF to learn the geometry of a transparent object, it must be able to ``see'' it from multiple camera views.  If the transparent object is not visible from any views, then it will have no effect on the loss function used in training, and thus not be learned.  We thus look for a way to improve visibility of transparent objects to NeRF.

One property that transparent objects share (e.g., glass, clear plastic) is that they are glossy and thus produce specular reflections when the camera view direction is opposite to the surface normal of the incident direction of light.  To the NeRF model, a specular reflection viewed from multiple angles will appear as a white point on a solid surface---i.e., $\mathbf{c} = [1, 1, 1]^T$ and $\sigma > 0$, while from other angles it will appear as $\sigma \le 0$.  As $\sigma$ is view-independent, NeRF learns a $\sigma$ between fully opaque and fully transparent for such points.

By placing additional lights in the scene, we create more angles from which cameras will see specular reflections from transparent objects---this results in NeRF learning a model that fills holes in the scene.  While the number and placement of lights for optimal training is dependent on both the expected object distribution and camera placement, in experiments (Sec.~\ref{sec:experiments_with_lights}) we show that increasing from 1 light to a 5x5 array of lights improves the quality of the learned geometry.

\section{Experiments}
\label{sec:experiments}
\vspace{-4pt}
\NEW{%
We experiment in both simulation and on a physical ABB YuMi robot.
We generate multiple datasets, where each dataset consists of images and associated camera transforms of one static scene containing one or more transparent objects.
% For simulation, the datasets are synthetic---rendered using Blender 2.92's physically-based path tracer with high quality settings.
% For physical datasets, we collect images using either an Intel RealSense or Cannon EOS 60D camera, but only use one camera type per dataset.
% To test grasps in simulation, we use a physics-based analysis of grasp wrench resistance.
% To test grasps in physical experiments, we have the ABB YuMI perform the grasp and lift the object.
}\OLD{%
To experiment with grasping transparent objects using NeRF and Dex-Net,\ken{Dex-NeRF} we generate multiple scenes in both simulation and real, and compute depth maps and grasps.}
We train NeRF models using a modified JaxNeRF~\cite{jaxnerf} implementation on 4 Nvidia V100 GPUs.  \NEW{We use an existing pre-trained Dex-Net model for grasp planning without modification or fine-tuning.  We can do this since NeRF models can be rendered to depth maps from arbitrary camera intrinsics and extrinsics, thus we match our NeRF rendering to the Dex-Net model instead of training a new one.} \OLD{Since we can render depth maps with arbitrary intrinsics and extrinsics given a trained model, we use an existing Dex-Net model trained with the intrinsics for PhoXi \yahav{RealSense?} \jeff{This should be associated with the Dex-Net model, which camera is it?} camera position 0.6\,m above the work surface.}

\subsection{\NEW{Datasets}}
\vspace{-4pt}
As existing NeRF datasets do not include transparent objects\NEW{, and existing transparent-object-grasping datasets do not include multiple camera angles,} we generate new datasets using 3 different methods: synthetic, Cannon EOS 60D camera with a Tamron Di II lens with a locked focal length, and an Intel RealSense. %  \yahav{Do we want to show the robot mounted?}. \jeffi{No, considering...}

For synthetic datasets, we use Blender 2.92's \NEW{physically-based} Cycles renderer with path tracing set to 10240 samples per pixel, and max light path bounces set to 1024.  We chose theses settings by increasing them until renderings were indistinguishable from the previous setting---finding that lower settings lead to dark regions and smaller specular reflections.  For glass materials, we set the index of refraction to 1.45 to match physical glass.
\NEW{We include 8 synthetic datasets of transparent objects: 2 scenes with clutter: light array and single light; 4 singulated objects from Dex-Net: Pipe Connector, Pawn, Turbine Housing, Mount; and 2 household objects: Wineglass upright and Wineglass on side. As these computationally demanding to render due to the high quality settings, we distribute these as part of the contribution.}

For the Cannon EOS \NEW{and RealSense real-world} datasets, we place ArUco markers in the scene to aid in camera pose recovery and take photos around the objects using a fixed ISO, f-stop, and focal length.  We use bundle adjustment from COLMAP~\cite{schoenberger2016sfm,schoenberger2016mvs} to refine the camera poses and intrinsics to high accuracy. \NEW{We include 8 physical datasets of transparent objects with a variety of camera poses: table with clutter, Dishwasher, Tape Dispenser, Wineglass on side, Flask, Safety Glasses, Bottle upright, Lion Figurine in clutter. The main difficulty in generating these datasets is calibration and computing high-precision camera poses.

The datasets, available at \url{https://sites.google.com/view/dex-nerf} differ from other datasets in their focus on scenes with transparent objects in a graspable setting, with over 70 camera poses each.%
}

\OLD{For the RealSense datasets, we mount the an Intel RealSense camera to a Fetch robot, using a 3D printed mount to keep the camera's pose relative to the arm fixed.  We then program the robot to have the camera trace an arc at multiple elevations, while recording the pose of the camera through forward kinematics.}

\subsection{Synthetic Grasping Experiments}
\vspace{-4pt}

\begin{figure}[t]
\begin{minipage}{\textwidth}
% CoRL text width is 5.5in
  \begin{minipage}[b]{3.15in}
    %\begin{figure}
      \centering
      \scriptsize
       \begin{tabular}{@{}c@{\hspace{0.05in}}c@{\hspace{0.05in}}c@{\hspace{0.05in}}c@{}}
       \includegraphics[width=0.75in,trim=300 200 300 400,clip]{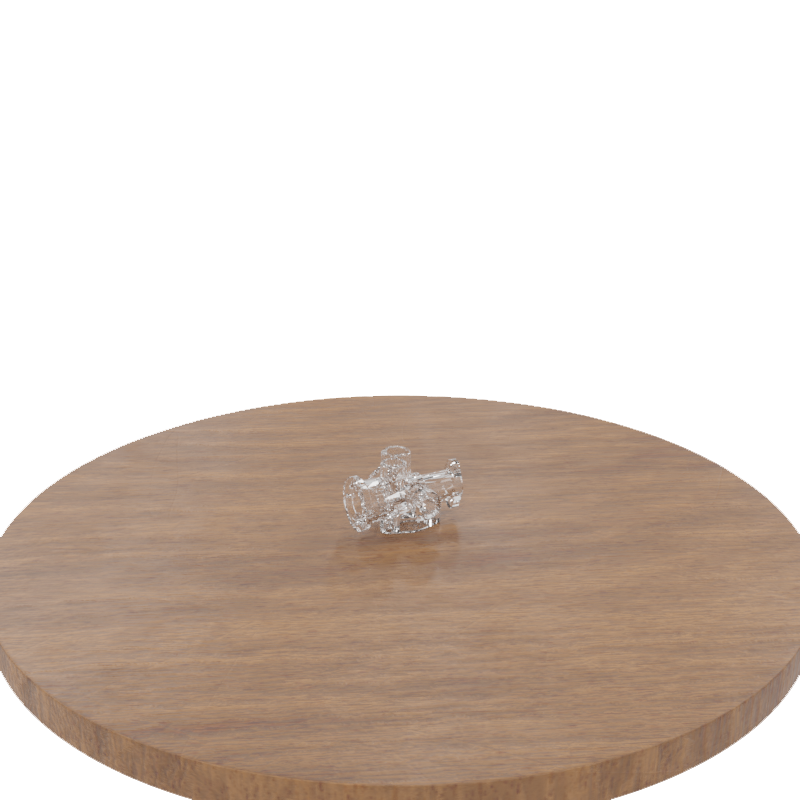} &
       \includegraphics[width=0.75in,trim=300 200 300 400,clip]{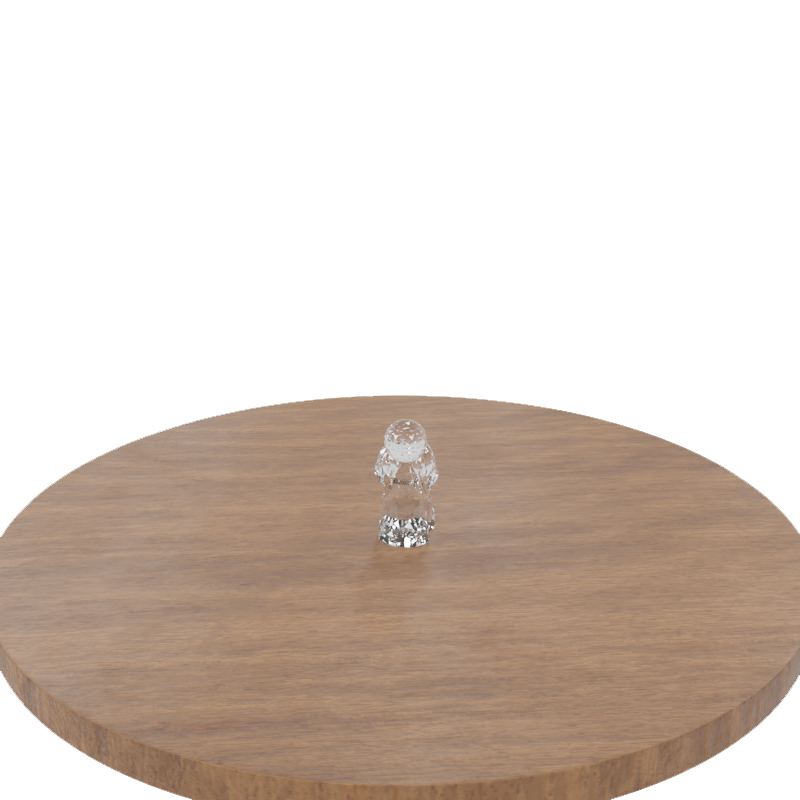} &
       \includegraphics[width=0.75in,trim=300 200 300 400,clip]{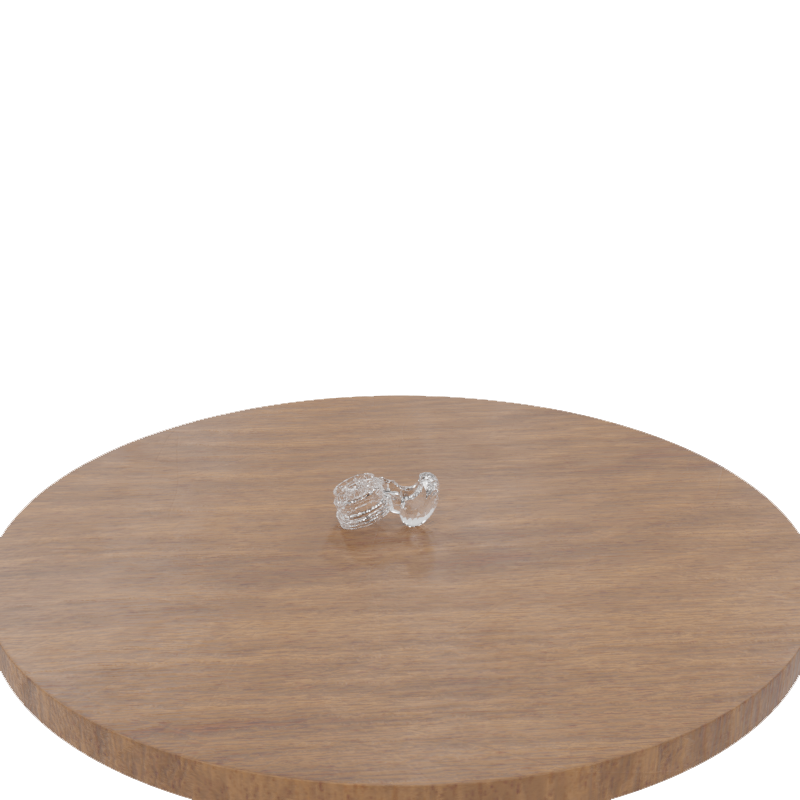} &
       %Pipe Connector & Pawn & Turbine Housing \\
       \includegraphics[width=0.75in,trim=300 200 300 400,clip]{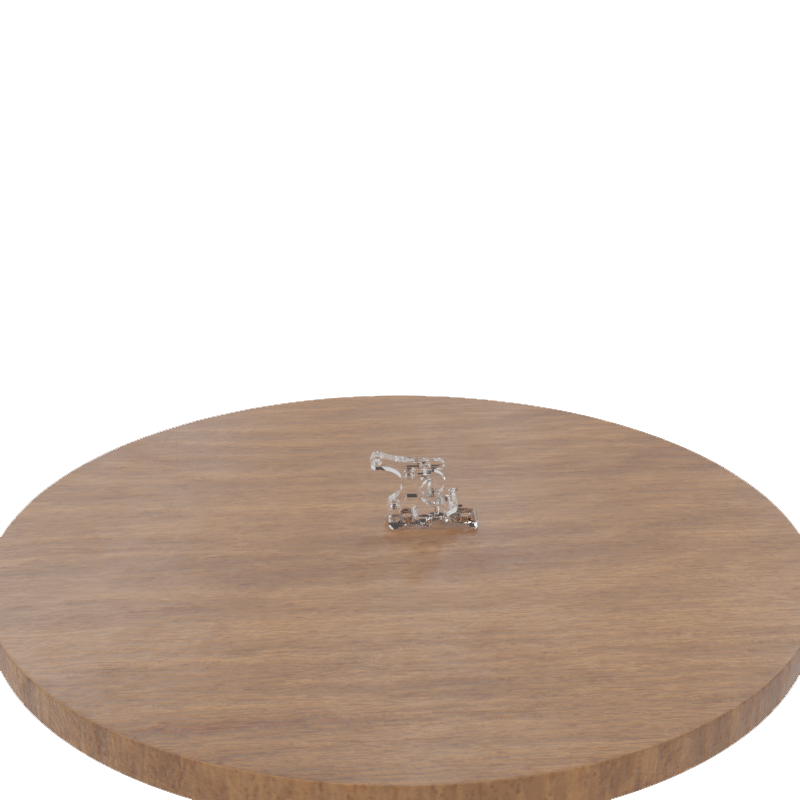} \\
       \includegraphics[width=0.75in,trim=394 404 414 456,clip]{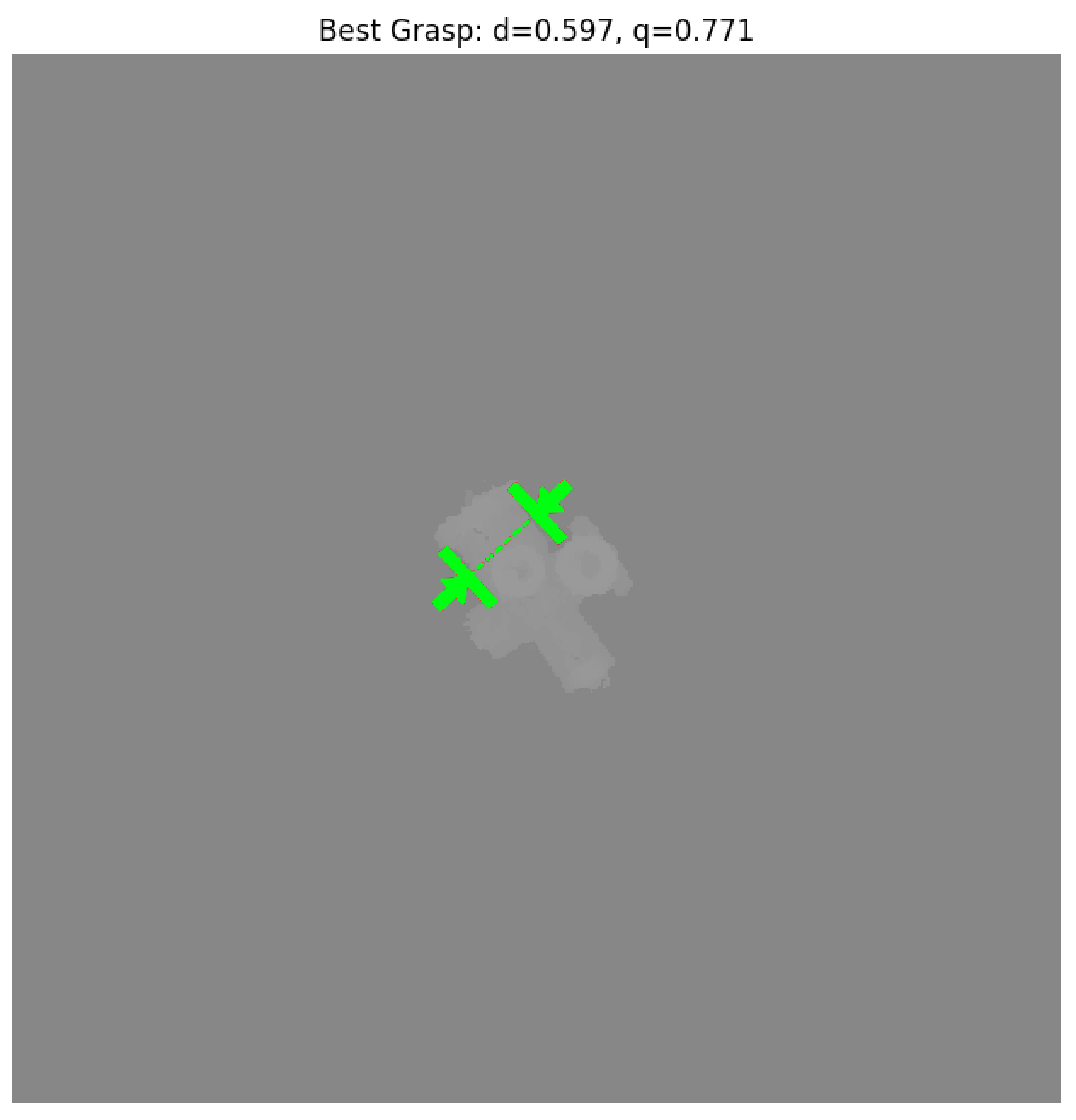} &
       \includegraphics[width=0.75in,trim=394 400 414 452,clip]{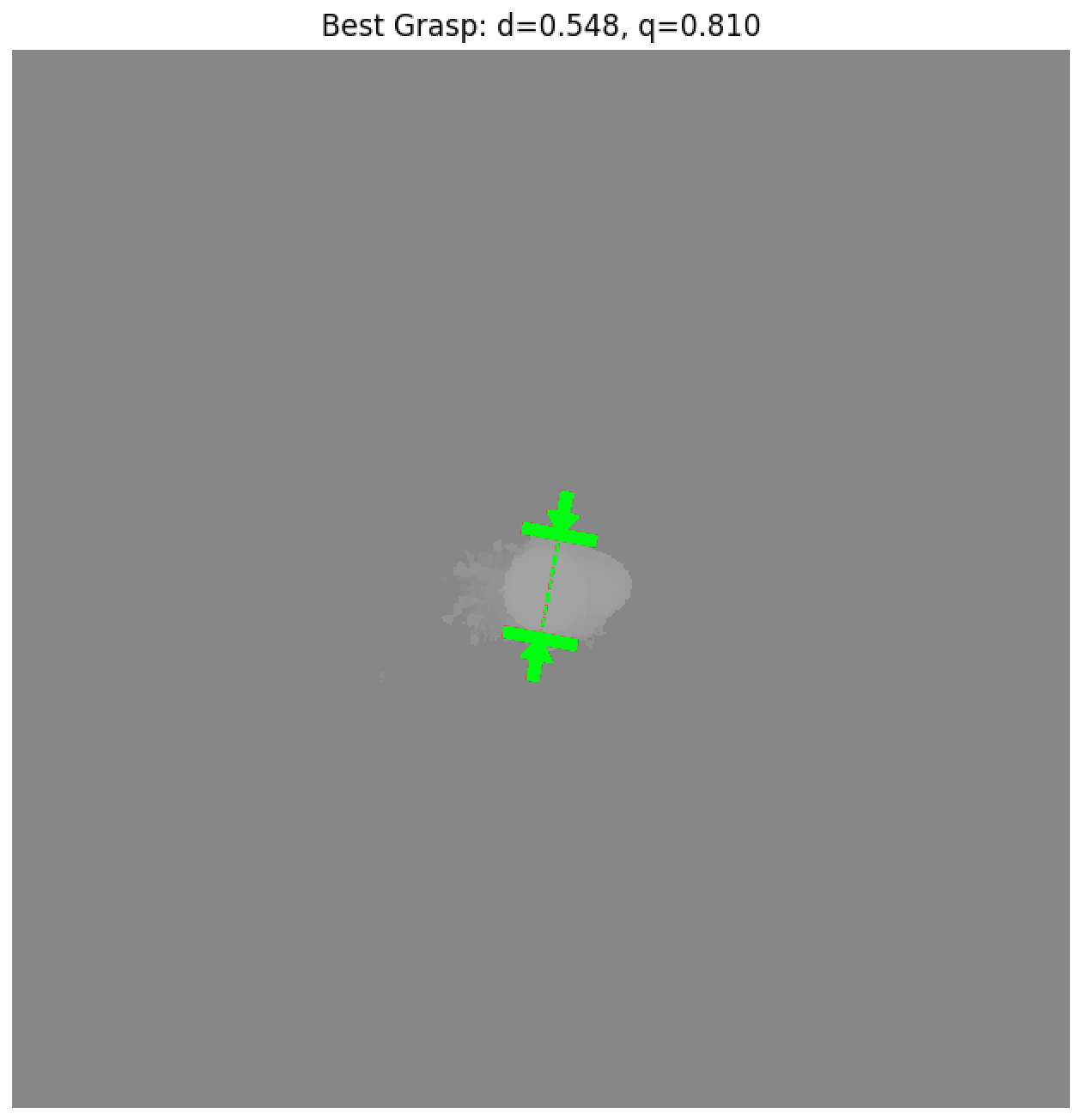} &
       \includegraphics[width=0.75in,trim=394 400 414 452,clip]{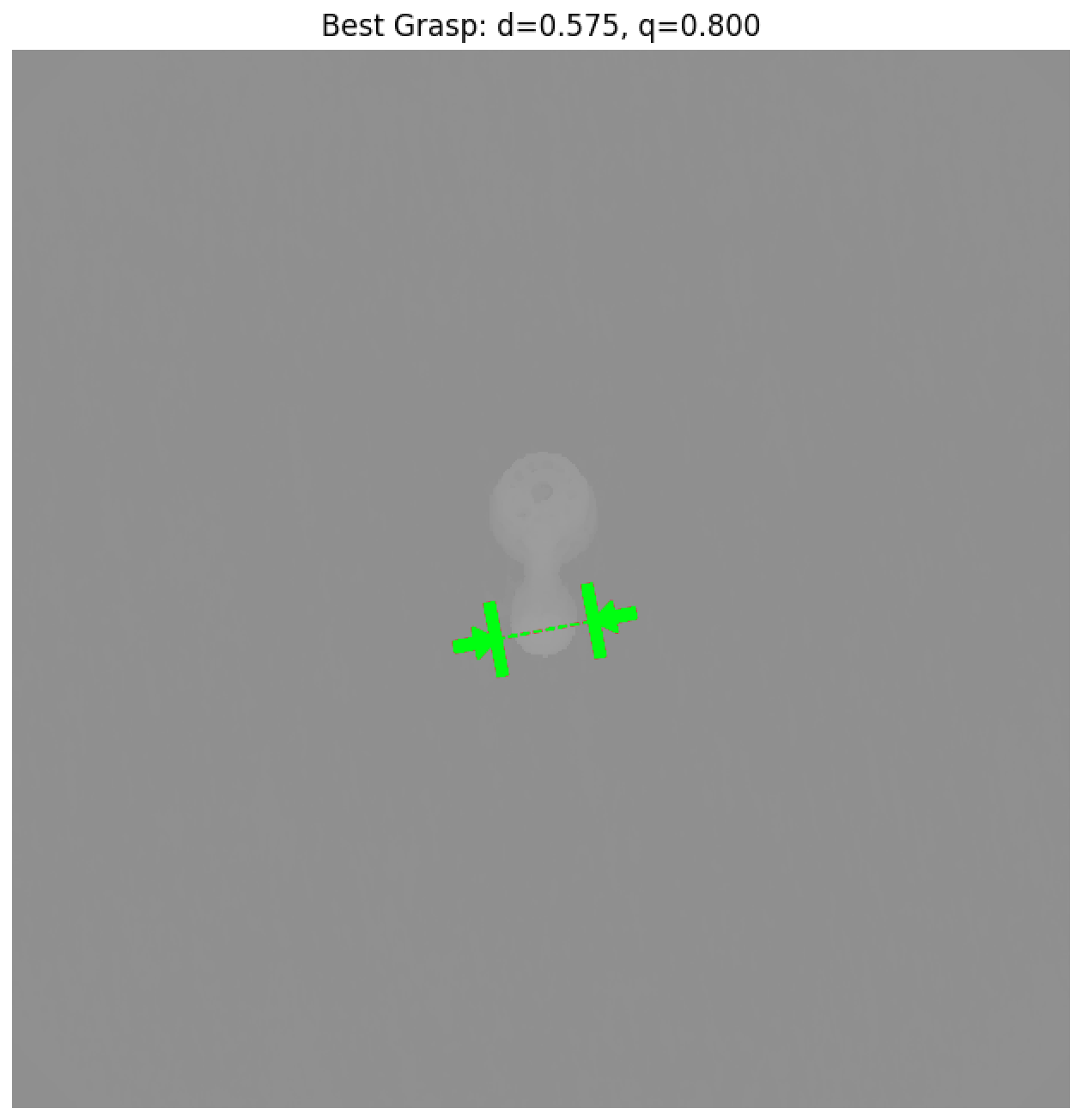} &
       \includegraphics[width=0.75in,trim=394 400 414 452,clip]{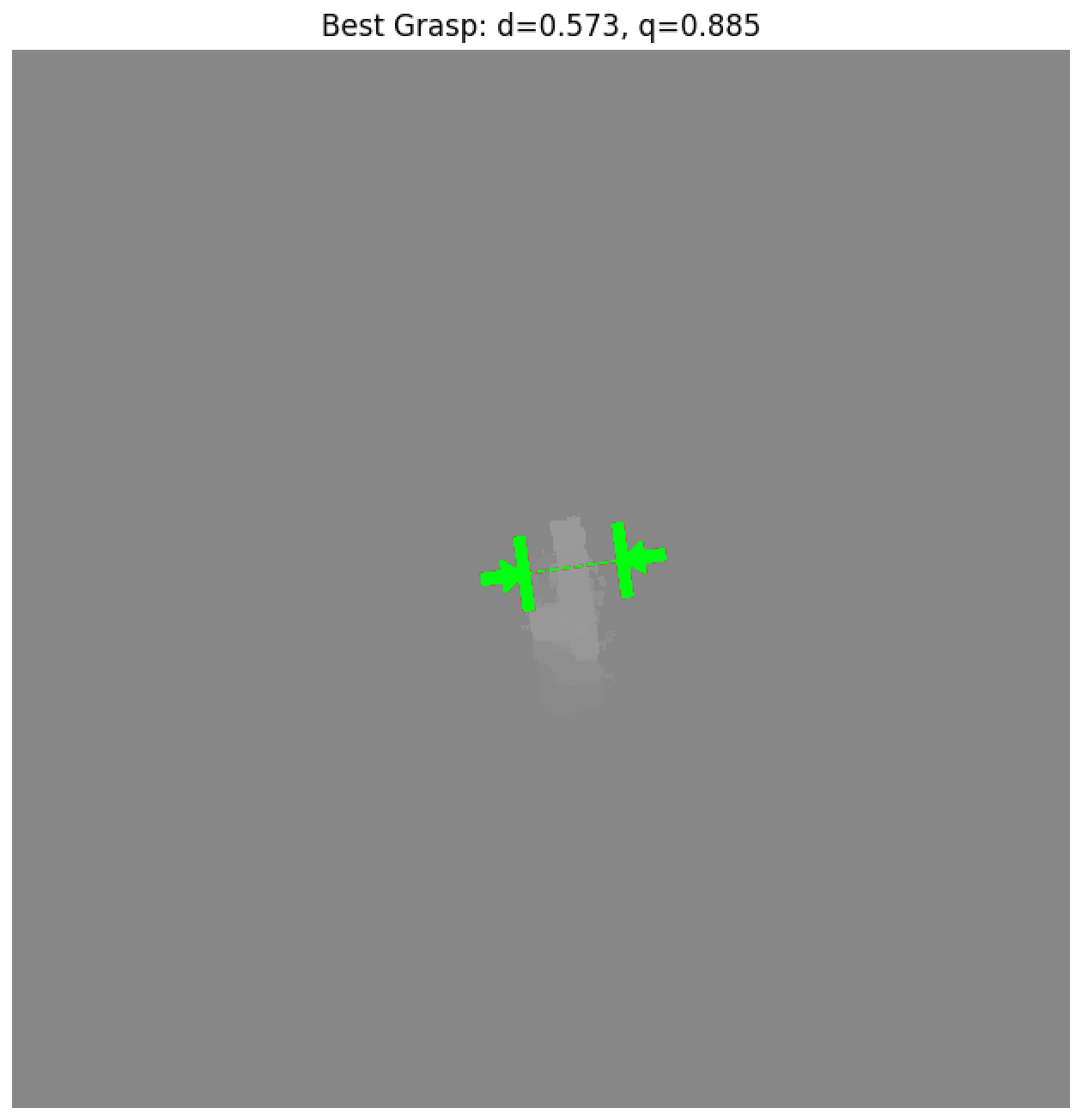} \\
       Pipe Connector & Pawn & Turbine Housing & Mount
    %   \includegraphics[width=0.75in,trim=300 280 300 320,clip]{images/grasps/wineglass_side.png} &
    %   \includegraphics[width=0.75in,trim=300 200 300 400,clip]{images/grasps/wineglass2_side.png} \\
    %   Mount & Wineglass (pose 1) & Wineglass (pose 2)
       \end{tabular}
       \captionof{figure}{\NEW{\footnotesize{\textbf{Synthetic singulated objects} used in simulation experiments.  \textbf{Top row:} image of the object in the training data.  \textbf{Bottom row:} computed depth map and candidate grasp.}}}
      \label{fig:synthetic_objects}
    %\end{figure}
  \end{minipage}\hfill%
  \begin{minipage}[b]{2.3in}
    %\begin{table}
    \centering
    \begin{tikzpicture}[
  inner frame sep=0,
  %every node/.style={font=\footnotesize},
  %background rectangle/.style={fill=olive!45}, show background rectangle,
  baseline=0pt]
\begin{axis}[
    ymin=0,
    ymax=100,
    xmin=0,
    xmax=100,
    %colorbrewer cycle list=Set1,
    ylabel={\footnotesize Grasp Success Rate (\%)},
    xlabel={\footnotesize Training epochs ($\times 1000$)},
    yticklabel style = {font=\scriptsize,xshift=0.5ex},
    xticklabel style = {font=\scriptsize,yshift=0.5ex},
    xtick={0,10,...,100},
    ytick={0,20,...,100},
    grid=both,
    grid style={line width=0.5pt, draw=gray!20},
    every tick/.style={draw=none},
    legend pos=south east,
    legend style={draw=none,fill=none, nodes={font=\scriptsize, inner sep=1pt}},
    legend cell align=left,
    width=2.45in,
    height=1.74in
]

\addplot+ [thick, mark options={scale=0.5}] coordinates {
(0,0)
(10,86)
(20,68)
(30,78)
(40,84)
(50,90)
(60,98)
(70,88)
(80,84)
(90,70)
(100,90)
};
\addlegendentry{Pipe Connector}
\addplot+ [thick, mark options={scale=0.5}] coordinates {
(0, 0)
(10, 4)
(20, 62)
(30, 70)
(40, 90)
(50, 82)
(60, 88)
(70, 82)
(80, 82)
(90, 82)
(100, 82)
};
\addlegendentry{Pawn}
\addplot+ [thick, mark options={scale=0.5}] coordinates {
(0,0)
(10,56)
(20,22)
(30,58)
(40,70)
(50,94)
(60,70)
(70,96)
(80,100)
(90,98)
(100,80)
};
\addlegendentry{Turbine Housing}
\addplot+ [thick, mark options={scale=0.5}] coordinates {
    (0, 0)
    (10, 0)
    (20, 0)
    (30, 73)
    (40, 54)
    (50, 97)
    (60, 95)
    (70, 92)
    (80, 89)
    (90, 97)
    (100, 94)
};
\addlegendentry{Mount}
\end{axis}
\end{tikzpicture}
    \captionof{figure}{\NEW{\footnotesize{\textbf{Grasp success rate vs training epochs.} As opposed to view-synthesis, which requires over 200k epochs, we observe high grasp success rates after 50k to 60k epochs.}}}
    \label{fig:grasps_vs_epochs}
  \end{minipage}
\end{minipage}
\end{figure}

% Singulated deer.  Relate to not needing priors

% Cluttered scene

% Opaque Scene
We also test the ability of \NEW{\algname{}}\OLD{the proposed system} to generate grasps on the synthetic singulated transparent Dex-Net object datasets.
\OLD{We also include a wineglass, that is not part of the Dex-Net dataset.  For this process, we (1) generate a scene by dropping the object in simulation to a stable pose, (2) render the images using Blender, (3) train the NeRF model, (4) compute the depth map, (5) run Dex-Net to compute and show the grasps and (6)}%
\NEW{For each dataset, we} evaluate the grasp in simulation using a wrench resistance metric measuring the ability of the grasp to resist gravity~\cite{mahler2018dex}.
\OLD{The results are shown in Fig.~\ref{fig:singulated_grasps}.}
\NEW{Fig.~\ref{fig:synthetic_objects} shows images of the synthetic objects, \algname{}-generated depth map, and an example sampled grasp for each.  To measure the effect of training time on grasp-success rate, we simulate and record grasps over the course of training.  In Fig.~\ref{fig:grasps_vs_epochs}, we observe that grasp success rate improves with training time, but plateaus between 80\,\% and 98\,\% success rate at around 50k to 60k iterations. This suggests that there may be a practical fixed iteration limit to obtain high grasp success rates}.  \OLD{As the photorealistic rendering of the images is a bottleneck, we curated the object set to include a diverse set of objects that are challenging to grasp.  In each case, the depth image the proposed system is able to generate allows Dex-Net to compute multiple candidate grasps.}

\NEW{We}\OLD{To} test \NEW{\algname{}}\OLD{the pipeline in a controlled setting \yahav{why controlled?}, we generate}\NEW{ on} a scene of a tabletop cluttered with transparent objects\OLD{, and run the pipeline shown in Fig.~\ref{fig:pipeline}}.
In this experiment, the goal is to grasp a transparent object placed in a stable pose \OLD{on a cluttered table, }in close proximity to other transparent objects.
The challenge is twofold: the depth rendering quality should be sufficient for both grasp planning and collision avoidance.
\OLD{To plan a parallel-jaw grasp from the rendered depth map we use a grasp planner based on Dex-Net 4.0 Grasp Quality CNN (GQ-CNN)~\cite{mahler2019learning} that samples grasp candidates from the image, evaluates each candidate and returns the grasp with the highest predicted quality along with its associated quality metric.}%
\NEW{Fig.~\ref{fig:pipeline} shows the}%
\OLD{The} robot and scene \OLD{are shown }in the upper left\OLD{ of Fig.~\ref{fig:pipeline}}, \NEW{and }the overhead image, depth, and computed grasp \OLD{are shown }inline in the pipeline, \NEW{and}\OLD{while} the final computed grasp with simulated execution is \OLD{show }in the upper middle image.
The final grasp contact point \OLD{computed by Dex-Net }was accurate to a 2\,mm tolerance\NEW{, suggesting that \algname{} with sufficient images taken from precisely-known camera locations may be practical in highly clutter environments}.%
\OLD{The simulated environment provides precise camera poses, but conversely is made more difficult due to the transparent objects having no imperfections.}

%To explore the use of NeRF for robotic manipulation of transparent objects, we incorporate the model in a grasp planning system. In this experiment, the goal is to grasp a transparent object placed in a stable pose on a cluttered table, in close proximity to other transparent objects. The challenge is twofold: the depth rendering quality should be sufficient for both grasp planning and collision avoidance. To plan a parallel-jaw grasp from the rendered depth map we use a grasp planner based on Dex-Net 4.0 Grasp Quality CNN (GQ-CNN)~\cite{mahler2019learning} that samples grasp candidates from the image, evaluates each candidate and returns the grasp with the highest predicted quality along wih its associated quality metric (see Fig.~\ref{fig:pipeline} and Fig.~\ref{fig:dishwasher}).

\subsection{\NEW{Physical Grasping Experiments}\OLD{Grasping real-world Objects}}
\vspace{-4pt}
\begin{figure}[t]
\begin{minipage}{\textwidth}
% CoRL text width is 5.5in
  \begin{minipage}[b]{2.35in}
    %\begin{figure}
      \centering
      \scriptsize
       \begin{tabular}{@{}c@{\hspace{0.05in}}c@{\hspace{0.05in}}c@{}}
       \includegraphics[width=0.75in]{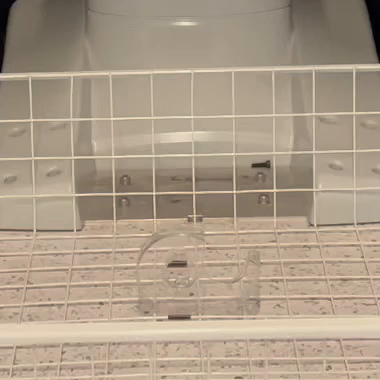} &
       \includegraphics[width=0.75in]{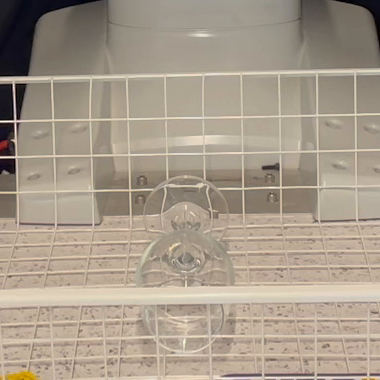} &
       \includegraphics[width=0.75in]{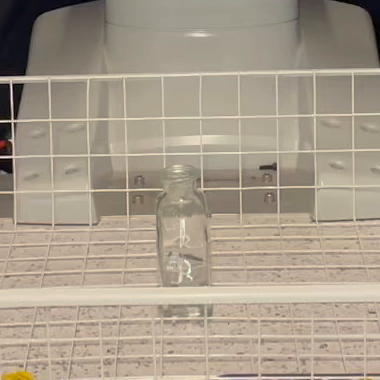} \\
       Tape Dispenser & Wineglass & Flask \\
       \includegraphics[width=0.75in]{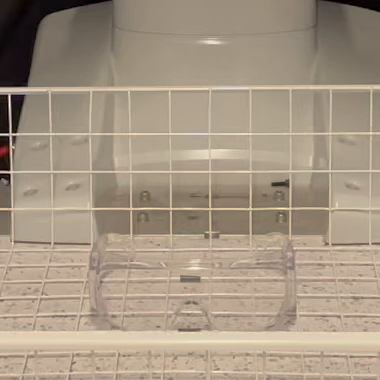} &
       \includegraphics[width=0.75in]{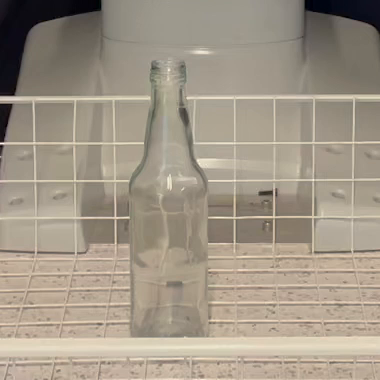} &
       \includegraphics[width=0.75in]{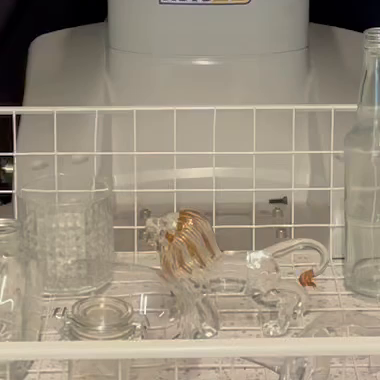} \\
       Safety Glasses & Bottle & Lion Figurine
       \end{tabular}
       \captionof{figure}{\NEW{\footnotesize{\textbf{Physical grasps objects.}  In the background is the base of the YuMi robot.}}}
      \label{fig:physical_objects}
    %\end{figure}
  \end{minipage}\hfill%
  \begin{minipage}[b]{3.1in}
    %\begin{table}
    \footnotesize
    \centering
    \begin{tabular}{lrrr} \toprule
          Object         &  PhoXi   &  Vanilla NeRF   & \algname{} \\
          \midrule
 Tape Dispenser &   0/10   &   0/10   & \bf  10/10   \\
 Wineglass      &   0/10   &   0/10   & \bf  9/10   \\
 Flask          &   0/10   &   1/10   & \bf  9/10   \\
 Safety Glasses &   0/10   &   0/10   & \bf  10/10   \\
 Bottle         &   0/10   & \bf 10/10  & \bf 10/10   \\
 Lion Figurine  &   0/10   &   3/10   & \bf  10/10   \\
 \bottomrule
    \end{tabular}\vspace{4pt}
    \captionof{table}{\NEW{\footnotesize{\textbf{Physical grasp success rate.}  For each object, we compute a depth map using a PhoXi camera, unmodified NeRF, and the proposed method for grasping transparent objects.  From the depth map, Dex-Net computes a 10 different grasps, and an ABB YuMi attemps the grasp.  Successful grasps lift the object.}}}
    \label{tab:physical_results}
    %\end{table}
  \end{minipage}
\end{minipage}
\vspace{-4pt}
\end{figure}

\NEW{%
To test the \algname{} in a physical setup, we place transparent singulated objects in front of an ABB YuMi robot, and have the robot perform the computed grasps.  We compare to 2 baselines: (1) \emph{PhoXi}, in which a PhoXi camera provides the depth map; and (2) \emph{Vanilla NeRF}, in which we use the original depth rendering from NeRF.
The PhoXi camera is normally able to generate high-precision depth maps for non-transparent objects.
All methods use the same pre-trained Dex-Net model, and both Vanilla NeRF and \algname{} use the same NeRF model---the only difference is the depth rendering.
We test with 6 objects (Fig.~\ref{fig:physical_objects}), and compute and execute 10 different grasps for each and record the success rate.  A grasp is successful if the robot lifts the object.  In Table~\ref{tab:physical_results}, we see that \algname{} gets 90\,\% and 100\,\% success rates for all objects, while the baselines get few successful grasps.  The PhoXi camera is unable to recover any meaningful geometry which causes Dex-Net predictions to fail.  The Vanilla NeRF depth maps often have unpredictable protrusions that result in Dex-Net generating unreliable grasps. %We also observe that physical grasp-success rate appears slightly higher than simulated grasp-success rates.  This may be due to the wrench-resistance analysis in simulation being overly pessimistic, the complexity of object geometries, and/or the presence of imperfections in physical objects that give NeRF more signal on which to train.
}

\OLD{We test in a real-world scenario of automated unloading the top-rack of a dishwasher loaded with glasses, jars, coffee presses, mug, and dishes.  We collect a dataset of 68 images using the Cannon EOS setup, and generate a depth map and grasps on objects in the rack.  See Fig.~\ref{fig:dishwasher} for the real image and depth image with an example associated suction grasp point computed by Dex-Net.  From the figure, we can see that the depth rendering includes all glass dishes, even though many are hard to see in the real image.  We also show a candidate suction grasp computed by Dex-Net~3.0~\cite{mahler2018dex} that places a suction contact on bottom of the upside-down french press.}

\subsection{Comparison to RealSense Depth}
\vspace{-4pt}
We qualitatively compare the the rendered depth map of the proposed pipeline against a readily-available depth sensor on scenes with transparent objects in real-world settings.  We select the Intel RealSense as it is common to robotics applications, readily available, and high-performance.  The RealSense, like most stereo depth cameras, struggles with transparent objects as they are unable to compute a stereo disparity between pixels from different cameras when the pixels are specular reflections or the color of the object behind the transparent object.  The RealSense optionally projects a structured light pattern on the the scene to aid in computing depth from textureless surfaces, however, in experiments we observed no qualitative difference with and without the light pattern emitter enabled.  To run this experiment, we use a Canon EOS as described in the main text, and use a RealSense to take a depth picture of the scene.  In this experiment, we observe that the RealSense is unable to compute the depth of most transparent objects, and often produces regions of unknown depth (shown in black) where transparent objects are.  On the other hand, the proposed pipeline produces high-quality depth maps with only a few areas of noise.  The results of these experiments is shown in Fig.~\ref{fig:nerf_vs_realsense}.

\subsection{One vs Many Lights}
\label{sec:experiments_with_lights}
\begin{figure}[t]
    \centering
    \includegraphics[width=0.95\linewidth]{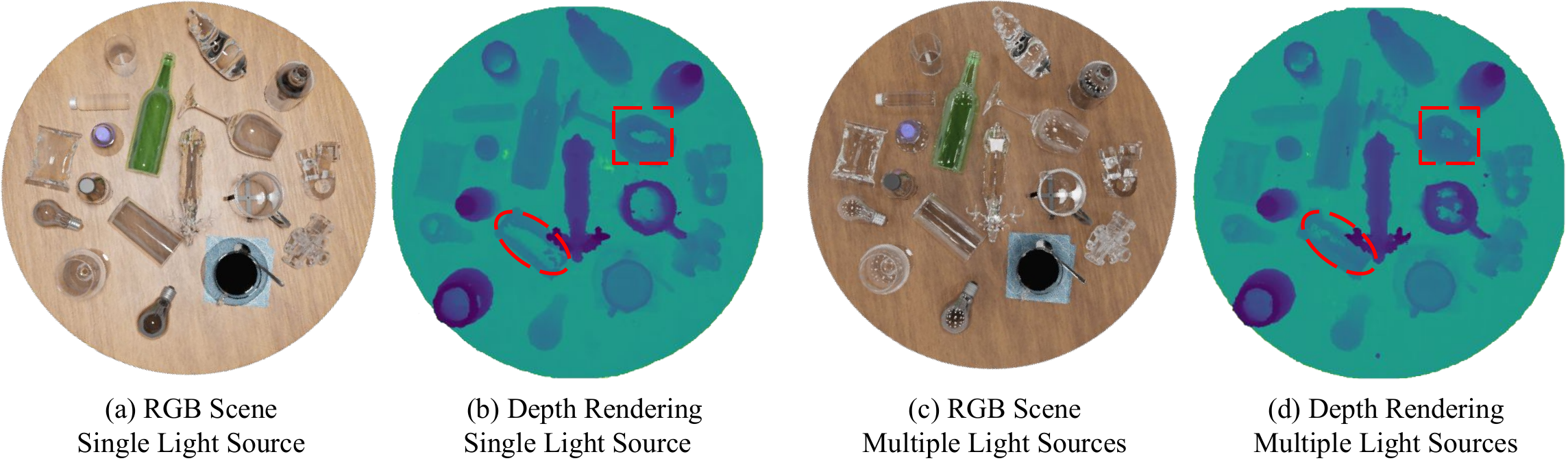}
    \caption{%\textbf{Specular reflections from additional light sources improve depth estimation.}
    \footnotesize{More lights mean more specular reflections, and result in better NeRF depth estimation of transparent surfaces.  In (a) and (b), we show a scene lit by a single overhead high-intensity light. In (c) and (d) we show the same scene lit by an overhead 5x5 array of lights.  The combined light wattage is equal in both scenes.  Images (a) and (c) are views of the scene, and (b) and (d) are the corresponding depth images obtained from the pipeline.  Two glasses on their sides are missing top surfaces (outlined in dashed red) in (b), while the effect is reduced in (d) due to the additional light sources.
    %The holes in the depth rendering are marked with red dots, and the holes are smaller when using more lights. 
    % \yahav{try to completely fill the holes and find a clearer way to point out the difference}
    }}
    \label{fig:single_multi_lights}
    \vspace{-4pt}
\end{figure}

\vspace{-4pt}
We experiment with different light setups to test the effect of specular reflections on the ability of NeRF to recover geometry of transparent objects.  We create two scenes (Fig.~\ref{fig:single_multi_lights}), one with a single bright light source directly above the work surface, and another with an array of 5x5 (25) lights above the work surface.  We set the total wattage of the lights in each scene to be the same.  Since most lights in the multiple light scene are further away from the work surface than the single light source, the scene appears darker, though more evenly illuminated.  The effect of the specular reflections is prominent on the lightbulb in the lower part of the image.  In the single light source, there is a single specular reflection, while in the multiple light scene, the reflection of the array of lights is visible. 

With the same camera setup for both scenes, we train NeRF models with the same number of iterations.  We show the depth rendering in Fig.~\ref{fig:single_multi_lights} and circle a glass and a wineglass on their side.  In the single-light source image, the closer surfaces of the glasses are missing, while in the multiple-light source depth image, the glasses are nearly fully recovered.  This suggests that additional lights in the scene can help NeRF recover the geometry of transparent objects better.

% \subsection{Ablation Study}
% \subsubsection{Depth Rendering Function}

% \subsubsection{Volume Density Threshold}

% We explore the...

% \subsection{Grasping}
% To explore the use of NeRF for robotic manipulation of transparent objects, we incorporate the model in a grasp planning system. In this experiment, the goal is to grasp a transparent object placed in a stable pose on a cluttered table, in close proximity to other transparent objects. The challenge is twofold: the depth rendering quality should be sufficient for both grasp planning and collision avoidance. To plan a parallel-jaw grasp from the rendered depth map we use a grasp planner based on Dex-Net 4.0 Grasp Quality CNN (GQ-CNN)~\cite{mahler2019learning} that samples grasp candidates from the image, evaluates each candidate and returns the grasp with the highest predicted quality along wih its associated quality metric (see Fig.~\ref{fig:pipeline} and Fig.~\ref{fig:dishwasher}).

\subsection{Workcell Setup}
\vspace{-4pt}
We experiment with a potential setup for a robot workcell in which a grid of overhead cameras captures views of the cluttered scene so that a robot manipulator arm can then perform tasks with transparent objects in the workcell.  We propose that a grid of overhead cameras would be practical to setup and would not obstruct manipulator tasks nor operator interventions.  The objective is to determine how many overhead cameras would be needed to recover a depth map of sufficient accuracy to perform maniplation tasks.

\begin{figure}[t]
    \centering
    \footnotesize
    \begin{tabular}{@{}c@{}c@{}c@{}c@{}c@{}}% trim=left bottom right top
    {\includegraphics[width=0.2\linewidth,trim=25 14 15 26,clip]{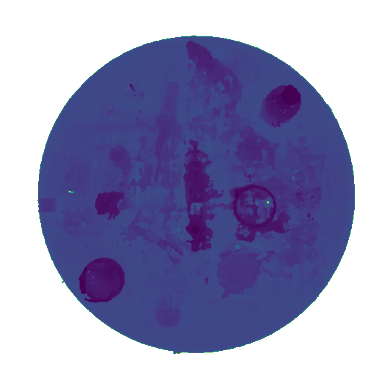}}& %
    {\includegraphics[width=0.2\linewidth,trim=25 14 15 26,clip]{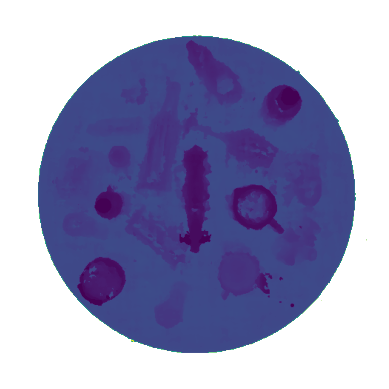}}& %
    {\includegraphics[width=0.2\linewidth,trim=25 14 15 26,clip]{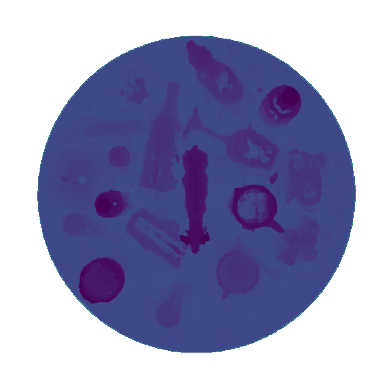}}& %
    {\includegraphics[width=0.2\linewidth,trim=25 14 15 26,clip]{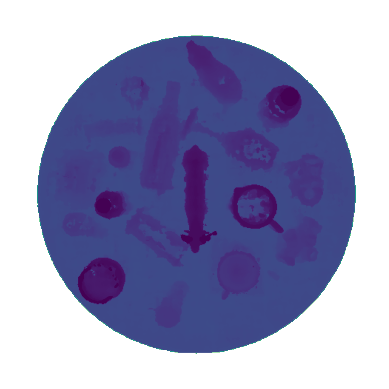}}& %
    {\includegraphics[width=0.2\linewidth,trim=25 14 15 26,clip]{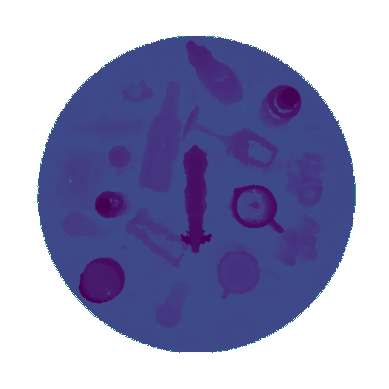}} \\
    $9$ Cameras&$16$ Cameras&$25$ Cameras&$36$ Cameras&$49$ Cameras
    \end{tabular}\vspace{-4pt}
    \caption{\footnotesize{{\bf Depth rendering using a grid of overhead cameras.} Using increasing amounts of overhead cameras improves the quality of the depth map and its utility in grasping, however, beyond a certain number of cameras there is a diminishing return.}} %
    \label{fig:camera_grids} %
    \vspace{-12pt}
\end{figure}

% \begin{table}[t]
% \centering
% \caption{Camera Grid for Robot Workcell}
% \scriptsize
% \begin{tabular}{@{}rrrr@{}}\toprule
%     & \multicolumn{1}{c}{train} & \multicolumn{2}{c}{test} \\
%      \cmidrule(lr){2-2} \cmidrule(l){3-4}
%      Cameras & \head{PSNR} & \head{PSNR} & \head{SSIM} \\
%      \midrule
%       4 (2x2) & 33.11 & 12.75 & 0.6913 \\
%       9 (3x3) & 30.99 & 25.54 & 0.8508 \\
%      16 (4x4) & 29.56 & 24.85 & 0.8090 \\
%      25 (5x5) & 29.23 & 26.06 & 0.8764 \\
%      36 (6x6) & 28.76 & 25.99 & 0.8169 \\
%      49 (7x7) & 28.85 & 28.34 & 0.8954 \\
%      \bottomrule
% \end{tabular}
% % \begin{tabular}{@{}lr@{}}\toprule
% % \multicolumn{1}{r}{Cameras} & 4 (2x2) \\
% % \midrule
% % train PSNR & 33.11 \\
% % test PSNR & 12.75 \\
% % test SSIM & 0.691 \\
% % \bottomrule
% % \end{tabular}
% \label{tab:camera_grid}
% \end{table}

% \begin{figure}
%     \centering
%     \begin{subfigure}[b]{0.16\textwidth}
%         \includegraphics[width=\textwidth]{images/cam_grid_9.png}
%         \caption{9 (3x3)}
%     \end{subfigure}
%     \caption{Caption}
%     \label{fig:my_label}
% \end{figure}

We place a 2\,m by 2\,m grid of cameras 1\,m above the work surface, and have them all point at the center of the work surface.  Each camera has the same intrinsics, and are evenly spaced along the grid.  We experiment with grids having 4, 9, 16, 25, 36, and 49 cameras.  The environment has the same 5x5 grid of lights as before.  For each camera grid, we train JaxNeRF for 50k iterations and compare performance.

After training, % from Table~\ref{tab:camera_grid},
we observe increasing peak signal to noise ratios (PSNR) and structural similarity (SSIM) scores with increasing number of cameras.  The 2x2 grid of cameras produces a high train-to-test ratio for PSNR, likely indicating overfitting to training data, and results in a depth map without apparent geometry. This ratio decreases with additional cameras.  The minimum number of cameras for this proposed setup appears to be around 9 (3x3) as its depth map is usable for grasp planning, while the 5x5 grid shows better PSNR and SSIM and ratio between train and test PSNR, and the 7x7 grid is the best.  See Fig.~\ref{fig:camera_grids} for a visual comparison.  Additionally, we trained 9x9, 11x11, and 13x13 grids, observing no statistically significant improvement beyond the 7x7 grid.

\section{Conclusion}
\vspace{-8pt}
In this work, we showed that NeRF can recover the geometry of transparent objects with sufficient accuracy to compute grasps for robot manipulation tasks. NeRF learns the density of all points in space, which corresponds to how much the view-dependent color of each point contributes to rays passing through it. With the key observation that specular reflections on transparent objects cause NeRF to learn a non-zero density, we recover the geometry of transparent objects through a combination of additional lights to create specular reflections and thresholding to find transparent points that are visible from some view directions.  With the geometry recovered, we pass it to a grasp planner, and show that the recovered geometry is sufficient to compute a grasp, and accurate enough to \OLD{perform the grasp} \NEW{achieve 90\,\% and 100\,\% grasp success rates in physical experiments on an ABB YuMi robot}.  %As NeRF was originally intendent for use with Lambertian surfaces, no dataset existed with transparent objects, we thus
We created synthetic and real datasets for experiments in transparent geometry recovery, but we believe these datasets may be of interest to researchers interested in extending NeRF capabilities in other ways and thus contribute them as part of this research project.  Finally, to test if NeRF could be used in a robot workcell, we experimented with grids of cameras facing a worksurface and their ability to recover geometry in potential setup, and showed the increased capabilities and point of diminishing return for additional cameras.

In future work, we hope to address one of the main drawbacks of NeRF---the long training time required to obtain a NeRF model.  Many research groups have sped up training time through improved implementations, new algorithms, new network architectures, pre-conditioned network weights, focused sampling, and more.  While these approaches apply to general NeRF training, we plan to exploit features specific to robot scenerios to speed up training, including using depth camera data as additional training data, manipulator-arm-mounted cameras to inspect regions of interest, and visio-spatial foresight to adapt to changes in the environment.
%TODO:? Similarly, we hope to improve accuracy 
\section*{Acknowledgments}{This research was performed at the AUTOLAB at UC Berkeley in
affiliation with the Berkeley AI Research (BAIR) Lab, Berkeley Deep
Drive (BDD), the Real-Time Intelligent Secure Execution (RISE) Lab, and
the CITRIS “People and Robots” (CPAR) Initiative. We thank our
colleagues who provided helpful feedback and suggestions, in particular
Matthew Tancik. This article solely reflects the
opinions and conclusions of its authors and do not reflect the views of the sponsors or their associated entities.}
\bibliography{references}  % .bib

\end{document}